\documentclass[journal]{vgtc}              
\onlineid{1414}
\vgtccategory{Research}
\vgtccategory{Research}

\vgtcpapertype{evaluation}

\graphicspath{{figs/}{figures/}{pictures/}{images/}{./}} 
\usepackage{tabu}                      
\usepackage{booktabs}                  
\usepackage{lipsum}                    
\usepackage{mwe}                       
\usepackage[pagebackref,bookmarks]{hyperref}
\usepackage{mathptmx}                  
\usepackage{textcomp}                  
\usepackage{times}                     
\usepackage{cite}                      
\usepackage{mathtools}
\usepackage{caption}
\usepackage{color}
\usepackage{enumitem}
\usepackage{amsmath}
\usepackage{amsfonts}
\usepackage{amssymb}
\usepackage{graphics}
\usepackage[utf8]{inputenc}
\usepackage{amsmath,bm}
\usepackage{multirow}
\usepackage{multicol}
\usepackage{subcaption}
\usepackage{tcolorbox}
\tcbuselibrary{breakable}
\title{\update{MV}-Bench: Benchmarking Multimodal Large Language Models for \update{Coordinated Multi-View} Interface Construction}
\author{%
  \authororcid{Yue Zhao*}{0000-0003-0365-5291},
  \authororcid{Hongxu Liu*}{0009-0003-6326-8848},
  \authororcid{Feiyu Wang}{0009-0008-0831-545X},
  \authororcid{Xiaoyu Yang}{0009-0008-3378-782X},
  \authororcid{Tong Ge}{0000-0002-6758-9652},\\
  \authororcid{Zhen Yang}{0000-0003-0670-4538},
  \authororcid{Chao Wang}{0009-0003-3555-7677}, and
  \authororcid{Qiong Zeng}{0000-0002-2827-8261}
}
\authorfooter{
  \item Yue Zhao is with Shandong Second Medical University. E-mail: zhaoyue@sdsmu.edu.cn
  \item Hongxu Liu, Feiyu Wang, Xiaoyu Yang, and Zhen Yang are with Shandong University. E-mail: liuhongxu\_0227@163.com, \\202515267@mail.sdu.edu.cn,  xiaoyuyang713@gmail.com, zhenyang@sdu.edu.cn.
  \item Tong Ge is with Bairong Inc.. E-mail: tgeconf@gmail.com.
  \item Chao Wang is with Ke Holdings Inc.. E-mail: chanceycn@gmail.com.
  \item * The authors contributed equally to this research.
  \item Qiong Zeng, the corresponding author, is with the School of Computer Science and Technology, Shandong University, and the Shandong Provincial Key Laboratory of Computing-Network Integration,  Shandong University. Email:  qiong.zn@sdu.edu.cn.
}
\abstract{%
Multimodal large language models (MLLMs) are increasingly expected to automate visualization development by generating code directly from visual designs. However, existing evaluations focus on single-chart generation, overlooking \update{coordinated multi-view interface construction}, a practically demanding task that requires jointly managing data semantics, multi-view coordination, and interactive logic. \update{As a result,} MLLMs' capabilities in this area \zq{remain} largely unexamined, while the field also lacks a dedicated benchmark for systematic assessment.
To address this gap, we introduce \update{MV-Bench, the first} benchmark for evaluating MLLMs on \update{coordinated multi-view interface construction. Rather than} relying on incomplete or inconsistent open-source \update{implementations, we leverage} Tableau workbook files as a principled \update{source of ground-truth, as they explicitly} encode data bindings, visual mappings, and interactions. We develop a multi-stage pipeline that converts these specifications into executable web-based interfaces via structured intermediate representations. Using this pipeline, we construct 92 base interfaces and systematically recombine them across chart types, datasets, and interaction patterns, \update{producing 1,048 verified benchmark instances} with executable \zq{code}, rendered \update{interfaces}, datasets, and interaction annotations. \update{Using MV-Bench,} we evaluate \update{five} state-of-the-art MLLMs, \zq{queried under a single-pass setting}, across \update{metrics measuring} visual fidelity, data binding correctness, and interaction completeness. While \update{the strongest} model achieves 75.45\% accuracy in \update{visual} layout reproduction, performance drops sharply to 21.71\% on data binding and 11.68\% on interaction completeness. These \update{results demonstrate} that current MLLMs can \update{reproduce} visual appearance but \update{remain fundamentally limited in generating} the data semantics and interactive logic \zq{required by coordinated multi-view interfaces. Further experiments with iterative refinement improve code executability but do not substantially narrow the gap in data binding and interaction generation}. 
}

\teaser{
  \centering 
  \includegraphics[width=\linewidth, alt={}]{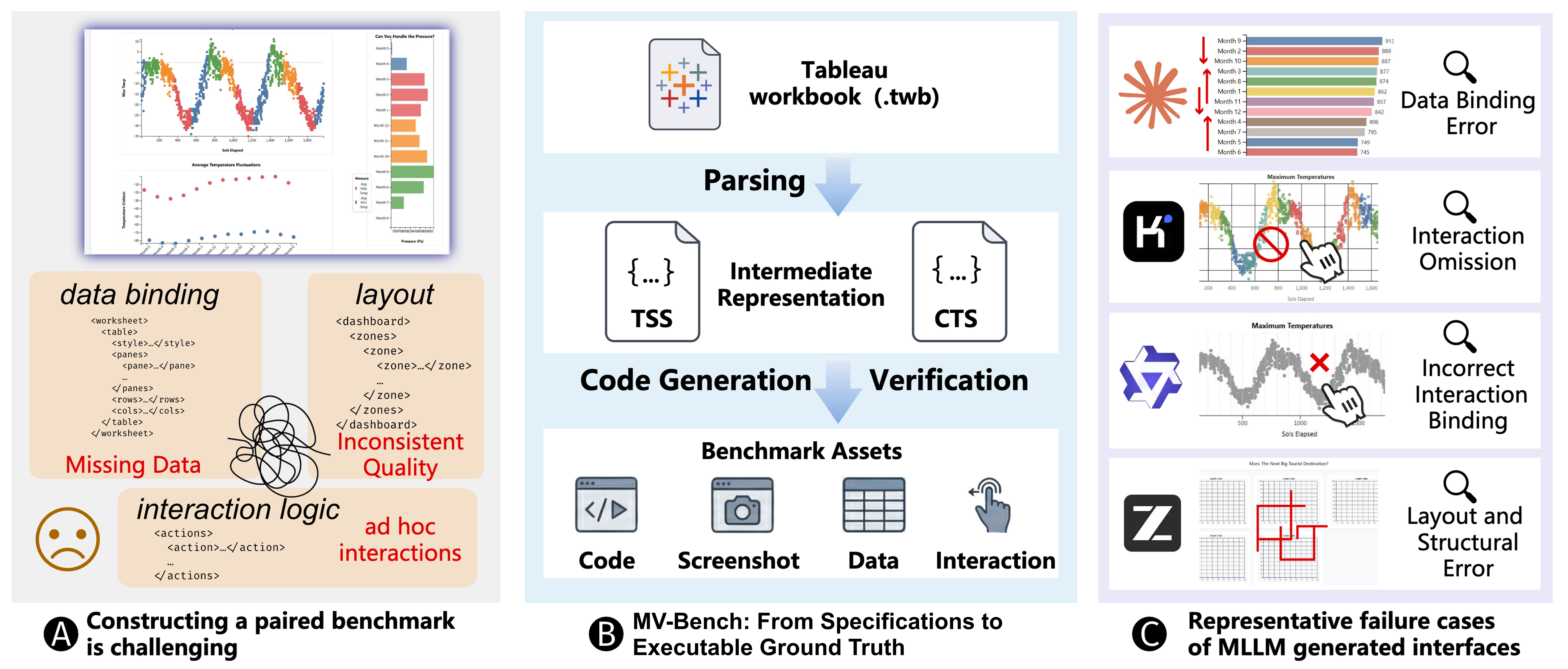}
  \caption{\update{MV-Bench} benchmarks multimodal code generation for \update{coordinated multi-view interface construction}.
  (a) Constructing interfaces from visual designs requires recovering data bindings, layout structures, and interaction logic from heterogeneous and often incomplete specifications.
  (b) \update{MV-Bench} transforms these fragmented workbook specifications into structured intermediate representations, and generates executable benchmark assets with aligned code, data, and interaction specification.
  (c) Using this benchmark, we systematically evaluate MLLMs and observe a consistent gap: while models can reproduce visual layouts, they frequently fail in data binding correctness, interaction implementation, and structural consistency.
  }
  \label{fig:teaser}
}

\newif\ifnotes
\notesfalse 

\newif\ifedited
 \editedtrue 
\editedfalse 

\definecolor{editcolor}{rgb}{0, 0, 0}
\definecolor{viseditcolor}{rgb}{1, 0, 0}
\definecolor{red}{rgb}{1,0,0}
\definecolor{darkgreen}{rgb}{0.11,0.6,0.24}
\definecolor{darkred}{rgb}{0.83,0.29,0.21}

\newcommand{\zq}[1]{{\color{black}#1}}
\newcommand{\update}[1]{\textcolor{black}{#1}}

\newcommand{\good}[1]{{\color{darkgreen}#1}}
\newcommand{\bad}[1]{{\color{darkred}#1}}
\newcommand{\worse}[1]{{\color{orange}#1}}

\def\etal{{\textit{et~al.}}}

\usepackage{soul}

\usepackage{afterpage}
\usepackage{xspace}
\usepackage[nameinlink,capitalise]{cleveref}
  \crefname{table}{Tab.}{Tabs.}
  \Crefname{table}{Table}{Tables}
  \crefname{section}{Sec.}{Secs.}
  \Crefname{Section}{Section}{Sections}
  \crefname{figure}{Fig.}{Figs.}
  \Crefname{Figure}{Fig.}{Figs.}

\begin{document}

\maketitle
\section{Introduction}
The rapid advancement of multimodal large language models (MLLMs) is reshaping how we think about visualization development~\cite{Dibia2023,Xie2024,Goswami2025,wang2025chartoptimiser,Niu2025}. Models such as GPT~\cite{OpenAI2024a} can interpret visual content and generate corresponding code with remarkable fluency, leading to a widespread assumption that visual design can be automatically translated into engineering implementation~\cite{Shen2022,Vaquez2024}. In visualization practice, design requirements are typically communicated through visual prototypes such as mockups, wireframes, or screenshots~\cite{Satyanarayan2017a,Liu2018,Ren2019,Ge2021}. 
Translating these visual prototypes into working code is therefore the core link between visual design and engineering implementation, and provides a natural and rigorous testbed for evaluating MLLMs' capabilities in visualization development.

However, existing image-to-code evaluations for visualization operate almost exclusively at the single-chart level~\cite{Han2023, Tian2025a, Yang2024a, Chen2021a, Masry2022a, Xu2024a}. In practice, visualization developers construct \update{coordinated multi-view interfaces}~\cite{Roberts2007a, Ward2010a, Heer2012, Munzner2015}, where real data drives visual encoding across multiple coordinated views connected through interactive mechanisms such as brushing and linking, filtering, and selection. By evaluating only at the single-chart level, existing work overlooks the data semantics, multi-view coordination, and interactive logic that \update{coordinated multi-view interface construction} demands, and therefore overestimates MLLMs' true capabilities in realistic visualization scenarios.
A key reason is the absence of a dedicated benchmark: without paired interface images and executable code with data and interaction, evaluation at this level is simply not possible.

Existing visualization benchmarks target chart-level tasks such as question answering and script generation~\cite{Masry2022a, Xu2024a, Yang2024a, Chen2021a}. Si~\etal~\cite{Si2024a} evaluate image-to-code translation for general web pages but without addressing data semantics and interaction logic. Lin~\etal~\cite{Lin2023} provide a large-scale corpus of real-world Tableau dashboards, but as structured specifications rather than image-code pairs. In short, existing resources each address part of the problem, but none provides a benchmark that combines interface-level image-code pairs with verifiable data and interaction ground truth. As a result, the three core pillars of \zq{coordinated multi-view interfaces}--data, visual representation, and interaction--remain untested in a unified evaluation setting.

Constructing such a paired benchmark is itself challenging: each ground-truth instance must bundle executable code with correct data bindings and verifiable interaction logic, a requirement that open-source visualization repositories, with their missing datasets, ad hoc interactions, and inconsistent quality, cannot reliably satisfy.
We observe that Lin~\etal's Tableau corpus~\cite{Lin2023}, while not directly providing image-code pairs, offers a principled starting point. Tableau workbook files (.twb) are self-contained, XML-based specifications that formally encode data bindings, visual mappings, multi-view layouts, and interaction definitions in a single artifact. 
However, these specifications cannot be directly executed or rendered as web-based interfaces, creating a gap between structured design prototypes and executable implementations.

We therefore develop a multi-stage pipeline that bridges the semantic gap between Tableau's authoring model and web-based rendering through structured intermediate representations. This pipeline systematically transforms high-level visualization specifications into executable interfaces while preserving data semantics and interaction logic. 
The first normalizes workbook semantics into an explicit structured format, and the second translates authoring-level descriptions into execution-level directives. A verification stage then checks the generated code against the original specification for consistency. Upon this pipeline, we build 92 base interfaces spanning five chart types and three core interaction patterns (brushing and linking, filtering, and selection and highlighting), each paired with executable code, rendered screenshots, underlying datasets, and structured interaction annotations. 
We systematically recombine these base interfaces across chart types, data sources, and interaction patterns, scaling \update{MV-Bench} to over 1,000 verified samples with controlled diversity and complete ground truth.

We evaluate \update{five} state-of-the-art MLLMs, including GLM~\cite{GLM2024}, Kimi~\cite{KimiTeam2025}, Qwen~\cite{Qwen2023}, Claude~\cite{Anthropic2024a}\update{, and GPT~\cite{OpenAI2024a}}, on image-to-code translation using \update{MV-Bench} across three \zq{metrics}: visual fidelity, data binding correctness, and interaction completeness. \zq{Each model is evaluated under a single-pass setting, producing its output in one pass from a fixed prompt} \zq{without iterative refinement to improve code executability}.
Our results reveal \update{a \zq{clear} gap}: \emph{while leading models achieve up to 75.45\% on visual layout reproduction, their performance drops sharply to 21.71\% on data binding correctness and 11.68\% on interaction completeness}. These findings indicate that, \zq{in this single-pass setting}, current MLLMs can reproduce the appearance of \update{coordinated multi-view interfaces} but fail to model the underlying data semantics and interaction logic that make them functionally correct.

In summary, we make the following contributions:
\begin{itemize}
\item \update{We identify a fundamental gap between single-chart generation and coordinated multi-view interface construction, and define interface-level image-to-code translation as a new evaluation task: given a reference screenshot, dataset, and interaction specification, generate executable code whose data bindings, multi-view layout, and interaction behavior match the reference.}
\item We introduce a pipeline that transforms structured visualization specifications into executable interfaces via intermediate representations, enabling the systematic construction of benchmarks with faithful data and interaction semantics.
\item We present \update{MV-Bench}, the first benchmark comprising over 1,000 verified \update{coordinated multi-view interfaces} with executable code, datasets, rendered outputs, and interaction annotations, supporting comprehensive evaluation beyond static visual appearance.
\item We provide the first systematic evaluation of MLLMs, \update{queried once per instance as coding agents}, on \zq{coordinated multi-view interface construction under a single-pass setting}. Our results reveal \zq{that} current models can reproduce visual layouts but struggle with data binding and interaction logic. \zq{Even with iterative refinement and execution feedback (\cref{sec:repair}), the gap remains, pointing to a real capability limitation.}
\end{itemize}
To facilitate future research, we have open-sourced our data and code at \url{https://osf.io/4w59m/overview?view_only=99ac625f654e47269f7114402870ad40}
\section{Related Work}
\subsection{Visualization Code Generation}
Generating visualization code from high-level specifications or intents  has been a long-standing goal in visualization. Existing efforts can be broadly categorized into three directions: specification and authoring ecosystems, learning-based approaches, and LLM-driven methods.

\textbf{Visualization specification and authoring ecosystems} form the foundation of programmatic visualization authoring. These systems can be further divided into \emph{grammar-based} and \emph{API-based} approaches. Grammar-based languages such as Vega-Lite~\cite{Satyanarayan2017a} provide concise abstractions for specifying visualizations through declarative mappings between data and visual encodings, with Satyanarayan~\etal~\cite{Satyanarayan2017a} proposing a layered grammar that composes interactive views via selections and transformations. In contrast, API-based systems such as D3~\cite{Bostock2011a}, ECharts~\cite{LI2018136a}, and Plotly~\cite{Sievert2020a} offer more flexible, low-level control over data-driven rendering, exemplified by Bostock~\etal's~\cite{Bostock2011a} fine-grained DOM manipulation for highly customized visualizations. Related authoring abstractions, such as Data Illustrator~\cite{Liu2018}, Charticulator~\cite{Ren2019}, and Libra~\cite{zhao2025libra}, further expose structured design operators or reusable interaction abstractions between high-level specification and low-level code.
While these languages establish standardized specification formats that underpin many downstream generation systems, they still require substantial expertise for effective code authoring.

\textbf{Learning-based approaches} translate natural language or structured queries to visualization specifications. Early Text2Vis work relies on rule-based systems and machine learning models to translate user queries into visualization specifications, though it struggles with semantically intricate queries~\cite{Shi2025}. Subsequent work instead formulates this as a \emph{sequence-to-sequence translation} problem: Luo~\etal~\cite{Luo2021a} develop an end-to-end neural translation model alongside a benchmark for NL2VIS translation accuracy, and follow-up work incorporates constraint-aware generation mechanisms--visualization-aware optimizations and constraint-based inputs--to enforce validity during decoding~\cite{Luo2022a}.
Although these methods demonstrate the feasibility of visualization synthesis, they are typically limited to generating single-view charts from explicit textual descriptions, without supporting complex interactions and multi-view coordination.

\textbf{LLM-driven methods} have significantly expanded the capability of visualization code generation. These approaches fall into three categories. \emph{Decomposition-based methods} break the generation process into structured subtasks~\cite{Tian2025a}, e.g., Tian~\etal~\cite{Tian2025a} decompose generation into intent understanding, planning, and code synthesis stages. \emph{Constraint- and grounding-based methods} improve reliability by disentangling high-level intent from data transformation~\cite{Wang2024a} or constraining outputs via schema-aware prompting and example retrieval~\cite{Li2024a}.
\emph{Agentic and workflow-based systems} extend beyond single-chart generation by integrating LLMs into multi-step analytical pipelines, including interface generation~\cite{Chen2022a} and end-to-end visual analytics workflows. For example, Zhao~\etal~\cite{Zhao2025a, Zhao2025b} employ LLM-based agents to coordinate end-to-end visual analytics workflows involving data processing, visualization, and interaction\zq{, while Moreira~\etal~\cite{RevisionWorkflowLLM2025} use a dataflow-based model that incorporates user intent at multiple scopes for human-AI alignment in authoring urban visual analytics systems}. Despite this rapid progress, few works systematically evaluate the construction of \update{coordinated multi-view interfaces} involving data-semantic grounding, multi-view coordination, and cross-view interactions, which is precisely the gap our benchmark aims to address.

\subsection{Image-to-Code Translation}
Translating visual designs into code is a practical need spanning UI and visualization development. We organize existing work into two categories: UI-level code generation and cross-platform UI migration.

\textbf{UI-level image-to-code generation} has evolved from early neural approaches to modern MLLM-based pipelines. Beltramelli~\etal~\cite{Beltramelli2017a} first demonstrated end-to-end translation from UI screenshots to domain-specific markup, and Deka~\etal~\cite{Deka2017a} contribute a large-scale dataset pairing screenshots with view hierarchies for learning-based UI understanding at scale. Recent web-focused work includes Lauren\c{c}on~\etal's~\cite{Laurencon2024a} large screenshot-to-HTML corpus, Li~\etal's~\cite{Li2024b} evaluation of vision-language models for web prototyping, and Xie~\etal's~\cite{Xie2025a} MLLM pipelines mapping visual widgets to code. Si~\etal~\cite{Si2024a} propose the first real-world benchmark for converting visual designs into functional HTML code and evaluate MLLMs on it, but their work targets visual and layout fidelity of static pages without addressing data semantics or interactive behavior.

\textbf{Cross-platform UI migration} addresses translating existing interfaces across frameworks while preserving functional equivalence. Gao~\etal~\cite{Gao2024a} employ rule-based mappings between source and target component hierarchies, while Gong~\etal~\cite{Gong2024a} propose a translation pipeline that preserves structural and stylistic properties across UI frameworks. These methods highlight the difficulty of maintaining behavioral consistency during translation, a challenge that intensifies in \update{coordinated multi-view interfaces} where data bindings and cross-view interactions must also be preserved.

\subsection{\zq{Coordinated Multi-View} Interface Design and Benchmarking}
Our work connects \zq{coordinated multi-view} interfaces with LLM evaluation. We review both aspects to position our contributions.

\zq{\textbf{Foundations of multi-view interface design} rely on} a rich theoretical and practical \zq{body of work}. Keim~\etal~\cite{Keim2008a} define visual analytics as analytical reasoning facilitated by interactive visual interfaces, emphasizing the coupling between data, models, visual representations, and user interactions in support of open-ended, highly interactive analytical workflows. Roberts~\etal~\cite{Roberts2007a} formalize coordinated multiple views as a core design paradigm, where brushing and linking, filtering, and selection connect views into a coherent analytical workspace. On the empirical side, Lin~\etal~\cite{Lin2023} conduct large-scale mining of Tableau dashboard designs, extracting recurring patterns in layout composition, chart type usage, and encoding strategies. These theoretical and empirical findings \zq{characterize what a multi-view} interface must encompass and inform the design dimensions of our benchmark: data binding, visual specification, multi-view layout, and interaction logic.

\update{We deliberately scope our contribution against this background. In Keim~\etal's sense, a visual analytics system entails deep, often bidirectional integration of data, analytical models, and visual representations for open-ended exploratory reasoning. The corpus underlying \mbox{MV-Bench}--Tableau and DMiner dashboards~\cite{Lin2023}--instead consists of \emph{coordinated multi-view dashboards}: juxtaposed views linked through simple declarative mechanisms such as filtering, highlighting, and brushing-and-linking. Yet these dashboards are far from trivial to build, and they remain underexamined: getting one right means resolving data bindings, multi-view layout, and cross-view interaction logic together, none of which single-chart benchmarks test\zq{~\cite{Masry2022a, Chen2025a, xia2025chartx}}. We therefore position \mbox{MV-Bench} as a benchmark for \emph{coordinated multi-view interface construction}, a necessary sub-problem of visual analytics authoring, and do not claim our findings generalize to settings requiring deeper data-model-visualization integration or open-ended analytical reasoning.}

\textbf{Benchmarking for visualization and code generation} is an active but fragmented area. Chen~\etal~\cite{Chen2025a} propose a visualization generation benchmark emphasizing evaluation within a feasible solution space rather than against a single reference. Related evaluation work also considers how LLMs' design preferences diverge from established best practices, arguing that evaluation should consider design quality beyond correctness~\cite{Wang2025a}. In software engineering, Jimenez~\etal~\cite{Jimenez2024a} operationalize code generation evaluation via real repositories and test suites. However, none of these benchmarks target image-conditioned generation of \update{coordinated multi-view interfaces} or evaluate the interplay of visual fidelity, data binding correctness, and interaction completeness.
\begin{figure}[htbp]
  \centering
  \includegraphics[width=0.99\linewidth]{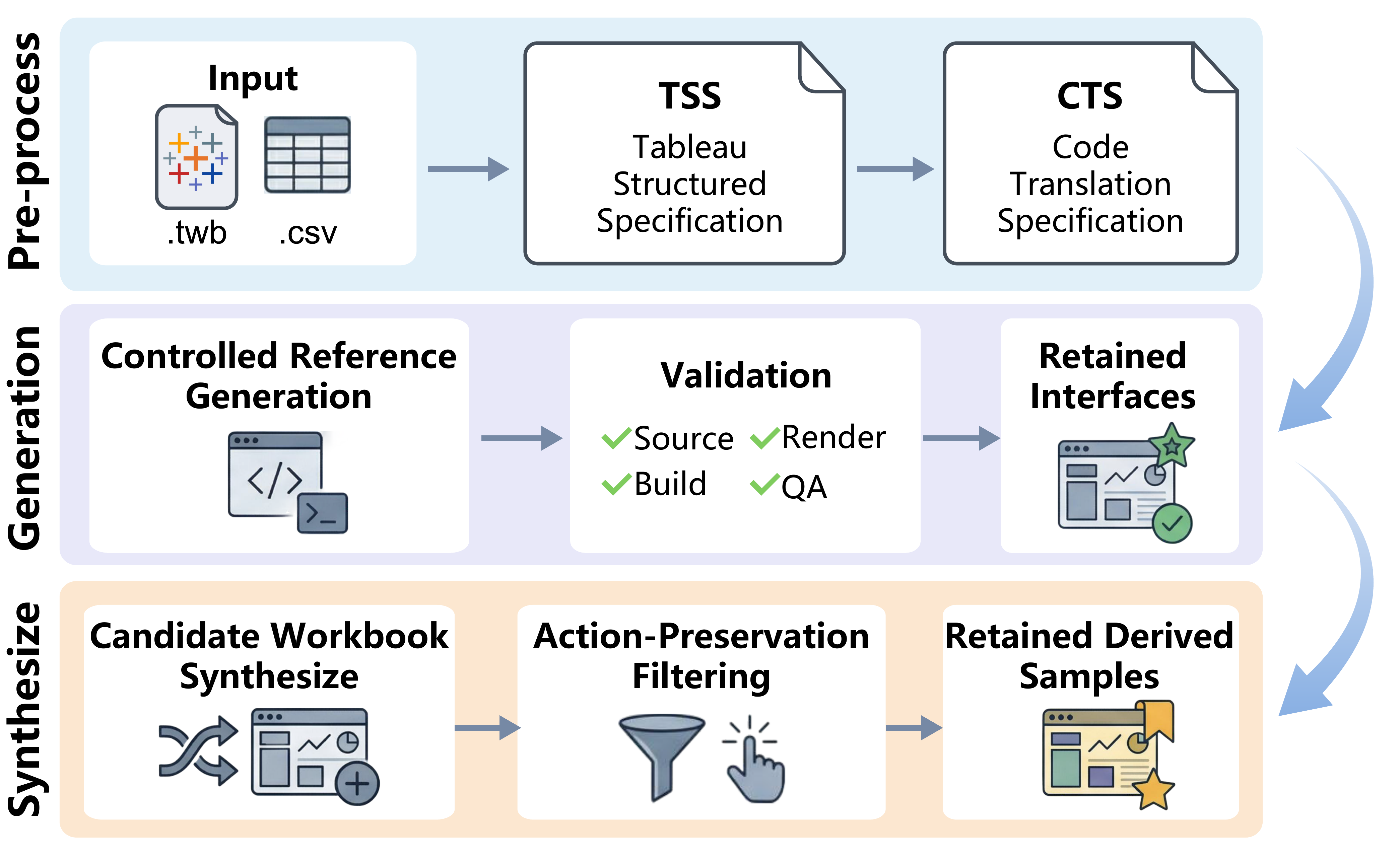}
  \caption{Construction pipeline of \update{MV-Bench}. We parse Tableau workbooks into structured intermediate representations--\emph{Tableau Structured Specification} (TSS) and \emph{Code Translation Specification} (CTS), generate executable web interfaces, and verify correctness at different levels. Derived samples are produced through controlled recomposition with interaction-preservation constraints.}
\label{fig:pipeline}
\end{figure}

\section{\update{MV-Bench} Construction}
\subsection{Overview}
We aim to construct a benchmark of executable \update{coordinated multi-view interfaces}, each paired with rendered screenshots, underlying data, and interaction annotations. Since Tableau workbooks already encode these four aspects in a structured format, we approach benchmark construction as a code translation problem, transforming workbooks from the DMiner corpus~\cite{Lin2023} into verified, executable web interfaces.

We address this through a pipeline of four stages (\cref{fig:pipeline}). We first collect and preprocess source workbooks and recover associated data assets (\cref{sec:curation}), then extract and translate each workbook into structured intermediate representations that bridge the semantic gap between Tableau's authoring model and web rendering (\cref{sec:parsing}). We generate executable reference code and verify its correctness through multi-level validation, yielding a set of verified base interfaces (\cref{sec:codegen}). Finally, to expand beyond these base interfaces, we recompose them into new combinations while preserving interaction semantics (\cref{sec:synthesis}).

\subsection{Data Preprocessing}
\label{sec:curation}
\subsubsection{Data source}
\label{sec:datasource}
We build on the DMiner corpus by Lin~\etal~\cite{Lin2023}, who crawled Tableau packaged workbooks (.twbx) from GitHub and curated 854 multi-view dashboards (2,990 views) for design rule mining. Each .twbx is a self-contained archive bundling a Tableau workbook (.twb) with its data extracts. Our corpus spans common chart types (\update{bar chart, line chart, scatterplot, pie chart, and data table}) and coordination patterns (\update{brushing and linking, filtering, and selection and highlighting}), and retains only dashboards with at least one cross-view interaction. \update{By construction, every one of the 1,048 benchmark instances therefore contains at least one interaction whose effect is observable in a non-source view; the supplemental material reports how this cross-view coordination is realized in practice, with selection and highlighting the dominant mechanism (84\% of interfaces), followed by filtering (6\%) and brushing and linking (1\%).} From our perspective, this corpus is valuable for three reasons:
\begin{itemize}
    \item Each .twb file encodes \zq{authoring specifications} relevant to our task in structured XML, including \zq{data} bindings, multi-view layout, and cross-view interactions.
    \item The corpus \zq{retains only dashboards with view coordination, so it} is enriched in the multi-view interfaces most relevant to our evaluation task.
    \item  The .twbx format bundles data extracts alongside the workbook, providing a starting point for data recovery.
\end{itemize}

At the same time, DMiner\zq{~\cite{Lin2023}} was not curated for executable benchmark construction, and not every workbook is directly usable in our setting. In particular, some workbooks depend on external or incomplete data sources. The preprocessing stage below therefore filters the source pool for workbooks whose data dependencies can be recovered and validated with reliability for downstream code generation and evaluation.

\subsubsection{Data recovery and cleaning}

DMiner~\cite{Lin2023} was constructed for design mining, which does not require complete access to the underlying data tables. Our benchmark, however, requires each retained workbook to support executable rendering and field-level verification: charts must be rendered from actual data tables, axes and legends must reflect recoverable fields, and cross-view interactions must be grounded in resolvable data attributes. We encounter two recurring data issues: some workbooks embed data inline within the .twb XML in Tableau-specific formats requiring extraction and conversion, and extracted data files contain format-level inconsistencies (CSV preamble rows, mixed character encodings, improperly escaped quotation marks). We address these with a recover-and-validate pipeline: for each workbook, we locate bundled extracts, inline tables, and referenced fields from the .twbx archive; normalize the data (removing preamble rows, standardizing to UTF-8, fixing quoting/escape errors); then perform a field-coverage check, retaining a workbook only if all fields referenced by its visual encodings and interaction definitions resolve in the recovered data. Workbooks with unrecoverable external sources, unparseable XML, or failed field-coverage are excluded. After this stage, 157 workbooks remain as the filtered source pool for the subsequent stages (Secs.~\ref{sec:parsing}--\ref{sec:codegen}); we treat these as a curated construction pool rather than a representative sample of the entire DMiner corpus, since the filtering biases retention toward dashboards with self-contained, machine-recoverable data dependencies.

\begin{figure}[t!]
  \centering
  \includegraphics[width=0.99\linewidth]{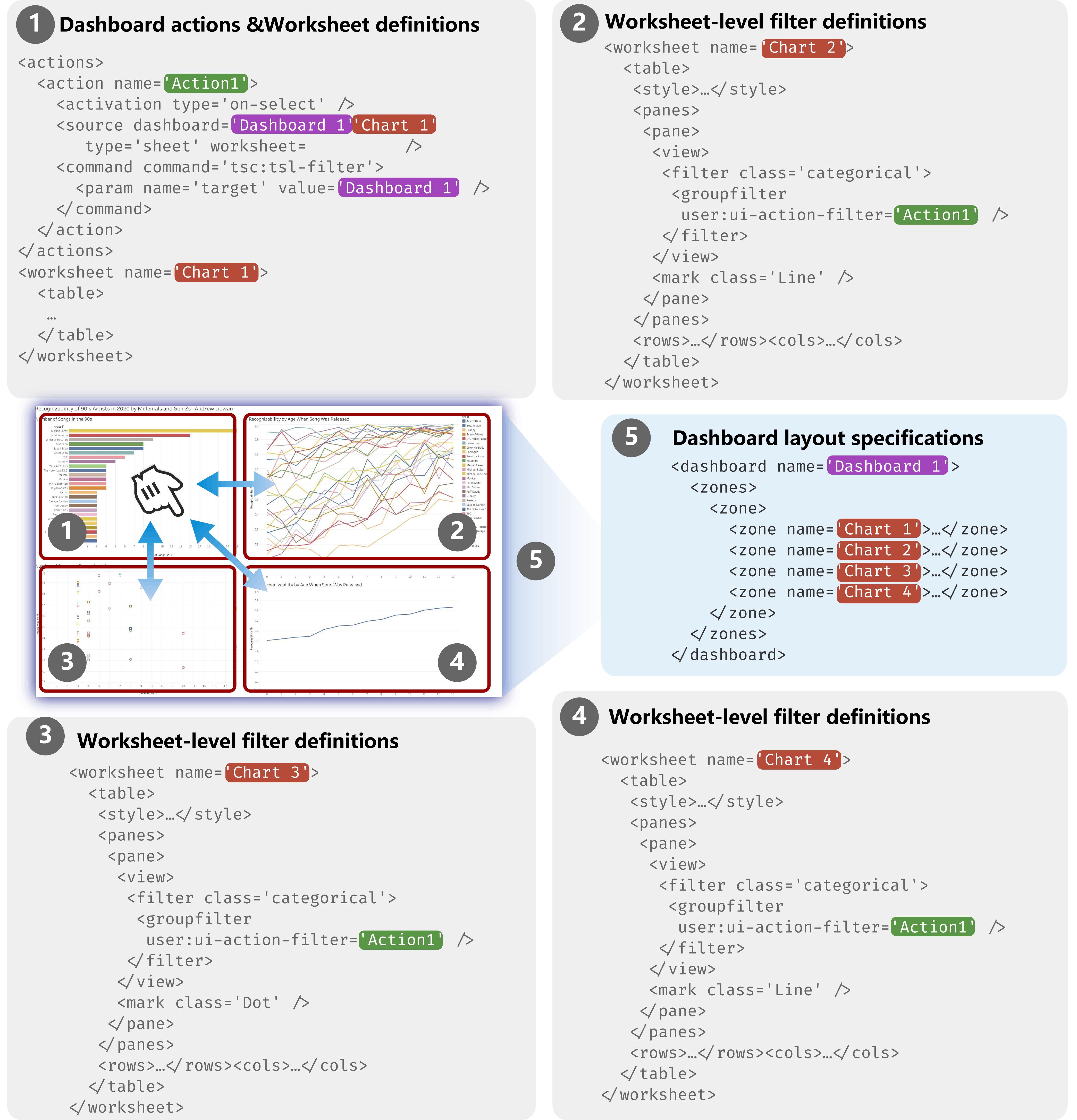}
  \caption{Example of distributed interaction semantics in a Tableau workbook. Cross-view interactions (e.g., filtering) are defined across dashboard actions, worksheet-level filters, and layout specifications rather than a single unified structure. \update{The numbered callouts trace one such interaction end-to-end: selecting a mark in \emph{Chart 1} (callout 1: dashboard action and worksheet definition) filters \emph{Charts 2--4} (callouts 2--4: worksheet-level filter definitions that reference the action) according to the zone layout declared in the dashboard specification (callout 5).}}
\label{fig:tableau}
\end{figure}

\subsection{Specification Extraction and Translation}
\label{sec:parsing}
To bridge the gap between Tableau's authoring format and executable web code, we extract and translate workbook content through two intermediate representations. We first describe the structure of Tableau workbooks that motivates this design, then detail each representation.

\subsubsection{Tableau Workbook Structure}
A Tableau workbook organizes content in two layers: \emph{worksheets} bind data fields to visual channels, set the chart type, and specify local filters and sort orders, while \emph{dashboards} compose worksheets into a coordinated interface by defining spatial layout and cross-view interactions such as filter and highlight actions. In the .twb XML, however, these layers are interleaved rather than cleanly separated, and as shown in~\cref{fig:tableau}, interaction and rendering behavior are distributed across dashboard actions, worksheet-level filter definitions, and layout specifications--making it hard to distinguish the author's design decisions from Tableau's internal defaults and rendering hints. We therefore decompose the translation into two steps: the \emph{Tableau Structured Specification} (TSS) normalizes what the workbook specifies as input (\cref{subsubsec:tss}), and the \emph{Code Translation Specification} (CTS) defines what the generated code must produce as output (\cref{subsubsec:cts}).

\subsubsection{Tableau Structured Specification}
\label{subsubsec:tss}
The TSS is a JSON-serialized representation that mirrors Tableau's own worksheet/dashboard architecture: at the \emph{view level} it records each worksheet's chart type, data-field-to-channel mappings, encoding parameters, and filtering conditions; at the \emph{dashboard level} it records spatial layout, cross-view coordination structures (filter actions, highlight bindings), and textual annotations. Crucially, the TSS preserves the author's design intent without committing to implementation decisions--e.g., it records that a field is placed on the columns shelf with a given sort order, but not whether this maps to an x-axis, a facet, or a grouping variable in the target framework.

\subsubsection{Code Translation Specification (CTS)}
\label{subsubsec:cts}
The TSS captures what the workbook specifies, but code generation also requires resolving rendering commitments left implicit at the workbook level (e.g., Tableau may imply stacking through shelf field order, or a filter action's scope through zone relationships rather than explicit declarations). The CTS translates these implicit semantics into concrete execution directives--chart intent, field-to-axis bindings, ordering and stacking constraints, legend placement, axis formatting, zone geometry, and interaction hooks with source-target-field triples. Where the TSS records ``field A is on the columns shelf,'' the CTS states ``field A must be rendered as the x-axis with ascending categorical ordering.'' We use the CTS as the primary grounding artifact for code generation, injecting sampled data rows into the prompt to expose column names, value types, and representative ranges, while the full datasets remain accessible through fixed runtime paths.

 \subsection{Reference Code Generation and Verification}
\label{sec:codegen}
Given the TSS and CTS produced in the previous stage, we now generate executable web interfaces and verify that they are correct enough to serve as benchmark ground truth. Implementation details including the technology stack and LLM configuration are provided in~\cref{subsec:details}.

\vspace{.3em}
\subsubsection{Generation} 
Since single-pass generation rarely produces correct multi-view interfaces, we adopt an iterative generate-and-repair strategy, \update{formulating} the generation input \update{as} the tuple $\textit{G} = \langle\textit{CTS}, \textit{TSS}, D_{\textit{sample}}\rangle$, where CTS provides the execution-level directives the code must satisfy, TSS the original authoring semantics as reference, and $D_{\textit{sample}}$ a small set of sampled data rows exposing column names, types, and representative value ranges. \zq{From this input, a}n LLM agent drafts the interface code--\emph{individual view components}, \emph{the coordinated multi-view layout}, and \emph{interaction handlers}. 
\zq{Each draft is checked by the same three-level verification used for final acceptance (source, build, and render validation): we compare it against CTS for missing views, unbound data fields, or omitted interaction hooks, and collect any compiler errors, runtime exceptions, and validation failures. A draft that passes all three levels is accepted and the loop terminates; otherwise these messages are fed back as feedback for the next attempt, for up to three attempts (\cref{subsec:details}). If none succeeds, the last is carried forward to verification and scored on its own merits. In the supplemental material, we detail this convergence behavior and how the same loop also serves as an iterative-repair evaluation.}
\begin{figure*}[t!]
  \centering
  \includegraphics[width=0.99\linewidth]{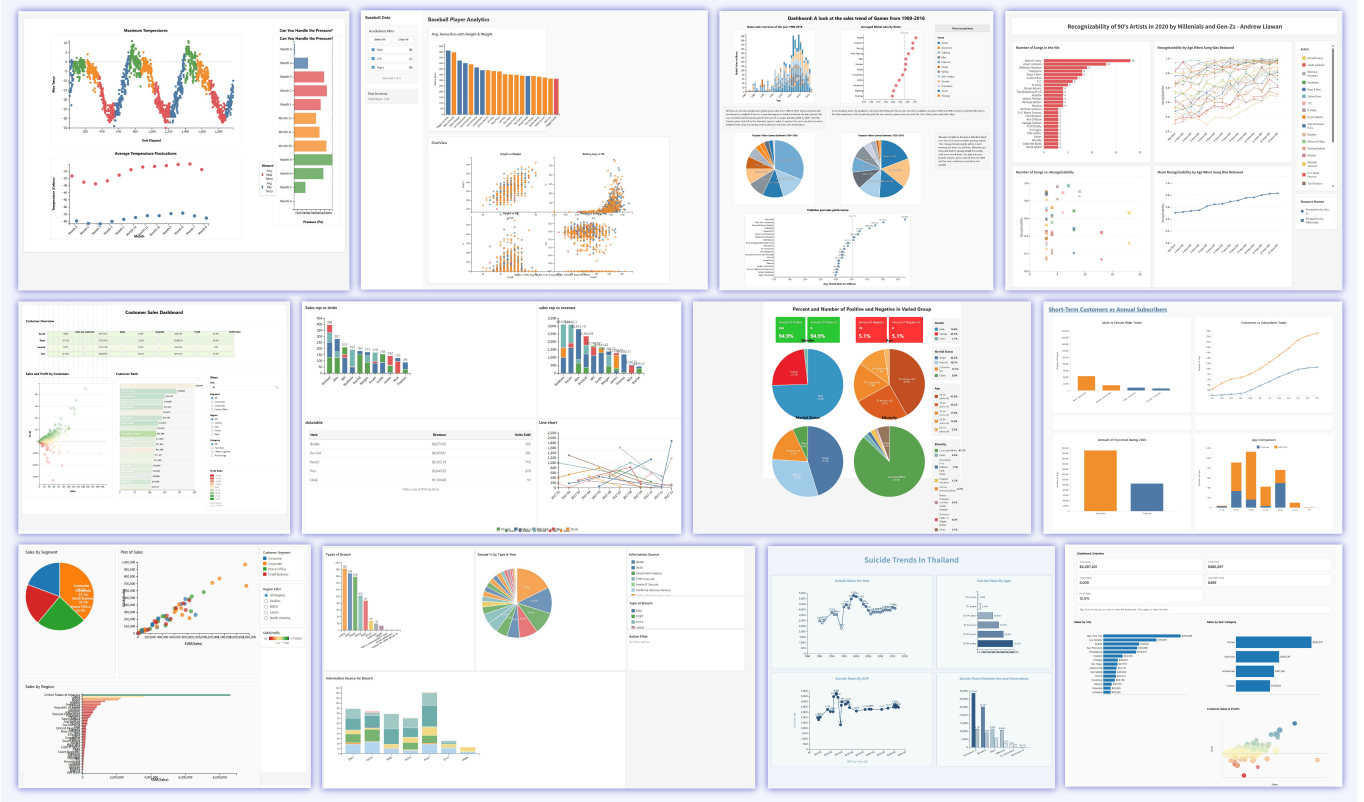}
  \caption{Overview of \update{MV-Bench}. The benchmark contains 92 base interfaces and 956 derived samples. Example dashboards illustrate variation in multi-view layout, chart combinations, and interaction patterns.}
\label{fig:benchmark-overview}
\end{figure*}

\vspace{.3em}
\subsubsection{Verification} 
After generation converges, we apply a final three-level verification to decide whether the result qualifies as benchmark ground truth. \emph{Source validation} checks that all data fields specified in the CTS are correctly referenced in the generated code and that the data file is properly loaded. \emph{Build validation} confirms that the project installs, passes lint and type checking, and builds without errors. \emph{Render validation} captures screenshots and verifies that all expected views are present and non-empty. We retain only candidates that pass all three levels, \zq{with source, build, and render validation passing at} 86.62\%, 67.52\%, and 58.66\%, respectively. In total, we obtain 92 base interfaces, each paired with executable source code, a rendered screenshot, the underlying dataset, and structured interaction annotations.

\subsection{Recomposition and Expansion}
\label{sec:synthesis}
The 92 base interfaces provide verified ground truth but cover a limited set of chart-type, data-source, and interaction-pattern combinations. To expand coverage, we synthesize additional samples by recomposing compatible components from existing base interfaces.

\subsubsection{Recomposition strategy} 
We construct new workbooks by combining layout structures, worksheet specifications, and data sources drawn from different base interfaces--e.g., a two-view bar+scatter dashboard may be recombined with a different dataset and a line chart replacing the scatter plot, preserving the original spatial arrangement and coordination pattern but with different visual content. Since different data sources may use different field names for similar concepts, we match fields across sources using heuristics based on data type, semantic role, and name similarity, then rewrite all internal workbook references to ensure consistency with the new data source.
 
\subsubsection{Interaction preservation}
Recomposition can break interaction semantics without affecting visual layout: a filter action propagating selections from a bar chart to a scatter plot relies on a shared data field that the new data source may lack, and brushing and linking similarly requires source and target views to share a common data dimension that may no longer hold after remapping. We address this with preservation rules: (1) excluding action-bearing worksheets from unconstrained remapping, (2) copying filter actions only when referenced fields exist and are type-compatible in the new data source, and (3) verifying that highlight bindings still resolve to valid targets. After applying these constraints, the synthesis pipeline produces 1,001 candidates, of which 956 pass both interaction-preservation filtering and the same three-level verification applied to base interfaces.

\subsubsection{Resulting benchmark.} The full \update{MV-Bench} comprises 1,048 instances: 92 base interfaces and 956 derived samples\zq{, covering the five chart types and three interaction patterns introduced in \cref{sec:datasource}}.
\cref{fig:benchmark-overview} illustrates representative instances from our benchmark. On average, each interface contains 4.4 coordinated views and 1.25 interaction actions. Since all instances are programmatically derived from Tableau dashboards, we assess fidelity against the original Tableau counterparts using standard visual similarity metrics (average SSIM 0.7042, MSE 0.0441, CLIP similarity 0.6770), which correspond to components of our $S_{static}$ measure defined below.


\subsection{Implementation Details}
\label{subsec:details}
\subsubsection{Parsing and generation}
We implement the Tableau parser in Python, extracting worksheet- and dashboard-level information and serializing it as TSS/CTS JSON files. We generate all reference implementations \zq{with GLM 4.7 as the code-generation agent (temperature 0, presence penalty 0.5), targeting a} React, TypeScript, and D3 stack for \zq{its} component-based architecture, static type checking, and fine-grained control over data-driven rendering and interactions\zq{; the per-component roles of this stack are detailed in \cref{subsubsec:architecture}}. The iterative repair loop (\cref{sec:codegen}) runs for up to 3 iterations per workbook, with automated error feedback at each step.

\zq{\subsubsection{Generated application architecture}
\label{subsubsec:architecture}
Every benchmark instance is generated from the same Vite project template—the reference implementation and each model-generated candidate differ only in the view, layout, and interaction logic the agent fills in. React function components organize each worksheet view and the dashboard layout, TypeScript provides static interfaces for the TSS/CTS-derived data contracts (chart configuration, field bindings, interaction hooks), and D3 handles data-driven rendering and binds interaction handlers within React's component lifecycle. The result is a self-contained, serverless single-page application: datasets are bundled as static assets at fixed relative paths, with no backend, database, or server-side rendering, and no configuration beyond installing pinned dependencies and starting the static build. This keeps every instance independently buildable, runnable, and replayable in the sandboxed Node.js environment described next.}

\subsubsection{Verification and evaluation environment}
We execute all generated code in a sandboxed Node.js environment with fixed dependency versions. For render validation, we use Puppeteer to capture screenshots at a fixed viewport of 1920$\times$1080. For interaction evaluation, we use Cypress to replay interaction episodes and verify post-state conditions through DOM inspection. We run the full pipeline on AutoDL Cloud. Code generation for the 92 base interfaces requires approximately 31 API hours. Recomposition and verification for derived samples takes an additional 255 hours.
\section{Evaluation Methodology}
We use \update{MV-Bench} to evaluate how well current MLLMs perform on \update{coordinated multi-view interface construction}. This section describes the evaluation setup, metrics, and evaluated models.

\subsection{Evaluation Setup}
\label{sec:setup}
We define a benchmark instance as \(b = (I_0, D, A, U)\), where \(I_0\) is the reference screenshot, \(D\) is the underlying dataset, \(A\) is the structured interaction specification, and \(U\) is the reference executable interface. Given \((I_0, D, A)\), a model \(f\) generates program code
\begin{equation}
P = f(I_0, D, A).
\end{equation}
\zq{Unless otherwise noted, we query $f$ once: each model produces $P$ from one fixed prompt in a single pass, with no feedback from execution and no repair. The reported scores therefore reflect what a model generates in a single attempt under a fixed prompt and target stack, not the best it could reach if allowed to iterate on execution feedback.} 
We execute \(P\) in a controlled sandbox to obtain a generated interface \(\hat{U}\), which is then compared against the reference interface \(U\). Since each interface may contain multiple coordinated views, our evaluation operates at two levels: the \emph{view level} for data binding correctness and the \emph{interface level} for static visual fidelity and interaction completeness. This design reflects the structure of \update{coordinated multi-view interfaces}, in which correct appearance alone is insufficient without correct data semantics and cross-view behavior.
\zq{To check whether single-pass scores are limited by the number of attempts rather than by the model's ability, we also evaluate all five models in a separate repair setting that reuses our reference pipeline's feedback signals (\cref{sec:repair}).}

\subsection{Evaluation Metrics}
\label{sec:metrics}
We first assess executability ($Exec.$) as a prerequisite gate, and then evaluate each generated interface \(\hat{U}\) against the reference \(U\) across three dimensions, corresponding to the core pillars of visual analytics identified by Keim~\etal~\cite{Keim2008a}: visual representation ($static$), data ($data$), and interaction ($int.$). We additionally report an aggregated overall score. All metrics are computed automatically from the rendered interface without manual inspection or code-level comparison.

\vspace{.3em}
\noindent\textbf{Executability} indicates whether the generated program builds and renders as a running interface. We define \(\mathrm{Exec}(\hat{U}) = 1\) if the program runs without fatal runtime errors, and 0 otherwise. All downstream metrics are computed only on executable outputs.

\vspace{.3em}
\noindent\textbf{Static visual fidelity}  measures whether \(\hat{U}\) reproduces the visual appearance of \(U\) in its initial state. we capture a screenshot \(\hat{I}_0\) under a fixed viewport (1920\(\times\)1080) and compute:
\begin{equation}
S_{static} = \frac{S_{CLIP} + S_{SSIM} + S_{nMSE} + S_{TreeBLEU}}{4},
\end{equation}
where all terms are normalized to [0, 1] with higher values indicating better agreement. $S_{CLIP}$ measures global visual-semantic similarity using CLIP~\cite{Radford2021}, capturing whether the generated interface conveys the same high-level visual content as the reference. $S_{SSIM}$ measures structural similarity via SSIM, focusing on spatial layout and luminance patterns. $S_{nMSE}$ is a normalized inverse MSE defined as $(1 - \mathrm{MSE}(\hat{I}_0, I_0) / \mathrm{MSE}_{\max})$, retaining sensitivity to pixel-level discrepancies that semantic metrics may miss.
$S_{TreeBLEU}$ evaluates component-level structural similarity by extracting a UI tree from each rendered page (traversing the DOM, collapsing library-specific wrapper nodes, and retaining semantically meaningful elements such as chart containers, axis groups, legends, and interactive controls), serializing it into a token sequence that preserves parent-child and sibling ordering, and computing a BLEU-style n-gram overlap against the reference sequence. \emph{We average the four terms with equal weight and report individual sub-scores alongside the aggregate}.

\vspace{.3em}
\noindent\textbf{Data binding correctness} measures whether \(\hat{U}\) renders the correct data values. For each view, we extract rendered data values from the DOM (e.g., axis tick labels, bar heights, data point coordinates, tooltip contents) and compare them against the corresponding values extracted from \(U\). We define:
\begin{equation}
S_{\text{data}} = \frac{1}{|V|} \sum_{v \in V} \mathrm{m}(v),
\end{equation}
where \(V\) is the set of views in the \update{coordinated multi-view interface} and \(\mathrm{m}(v)\) is the proportion of extracted data values in each view \(v\) that match the reference data. We first verify that the generated code loads the correct data file, then check whether the rendered values correspond to the expected fields and aggregations. A view that fails to load data or renders fabricated values receives \(\mathrm{m}(v) = 0\).

\vspace{.3em}
\noindent\textbf{Interaction completeness} measures whether \(\hat{U}\) reproduces the required behavioral effects under user actions. We compile the interaction specification \(A\) into replayable test episodes. Each episode \(\tau\) consists of: (1) a trigger action specifying the interaction type and target element (e.g., click on a data point in a scatter plot), (2) the expected affected views, and (3) post-state conditions describing the observable effect (e.g., corresponding bars in a linked bar chart become highlighted). We replay each episode from reset initial state and verify post-state conditions through DOM inspection. We define:
\begin{equation}
S_{\text{int.}} = \frac{1}{|T|} \sum_{\tau \in T} \mathrm{p}(\tau),
\end{equation}
where \(\mathrm{p}(\tau) = 1\) if all post-state conditions are satisfied within a timeout window, and 0 otherwise.
 
\vspace{.3em}
\noindent\textbf{Overall.} We first compute scores for each interface individually, then average across all interfaces so that each interface contributes equally regardless of its size or number of interaction episodes.
\begin{equation}
S_{\text{overall}} = \mathrm{Exec}(\hat{U}) \cdot \frac{1}{3} \left( S_{\text{static}} + S_{\text{data}} + S_{\text{int.}} \right).
\end{equation}
Non-executable outputs receive an overall score of zero. We weight the three dimensions equally as a baseline aggregation and report individual dimension scores to support analysis under alternative weightings.

\subsection{Evaluated Models}
\label{sec:models}
We evaluate five multimodal code-generation models: GLM 4.6V~\cite{GLM2024}, Kimi K2.5~\cite{KimiTeam2025}, Qwen3.5 Plus~\cite{Qwen2023}, Claude Sonnet 4.5~\cite{Anthropic2024a}, and GPT 5.4~\cite{OpenAI2024a}. \zq{We selected them by four criteria:
\begin{itemize}
\item \textbf{Multimodal input.} The model can jointly condition on a reference screenshot, tabular data, and a structured specification.
\item \textbf{Structured code generation.} The model generates code in the target stack our benchmark uses—React, TypeScript, and D3.
\item \textbf{Stable API access.} The model exposes a stable API compatible with our sandboxed batch-evaluation pipeline and its rate limits.
\item \textbf{Adoption.} The model is widely used, so the results are informative to practitioners and tool builders.
\end{itemize}
}
\update{We also attempted to evaluate Gemini~\cite{geminiteam2025} under the identical protocol, but excluded it from the main results because the third-party agentic coding-CLI tooling our batch-evaluation harness relies on did not reliably drive it, violating criterion (iii); the supplemental material reports this attempt and the resulting partial scores for transparency.}
\vspace{.5em}
\noindent\textbf{\zq{Query Protocol.}} \zq{We query all five models under an identical protocol to isolate model capability from prompting differences. Each model is invoked once per instance as a coding agent through the Claude Agent SDK\zq{~\cite{Anthropic2024a}}. The agent reads the inputs from disk and writes its implementation using a fixed instruction template that specifies the target stack (React, TypeScript, and D3), following the single-pass setting of \cref{sec:setup} with no feedback-based revision. The full prompt template, and any deterministic post-processing applied before evaluation, are described in the supplemental material; all code runs in the sandboxed environment of \cref{subsec:details}.
}

\begin{table*}[t!]
  \centering
  \footnotesize
  \setlength{\tabcolsep}{4.9pt}
  \renewcommand{\arraystretch}{1.25}
  \caption{Overall performance of the evaluated models on \update{MV-Bench}. Results are macro-averaged and reported separately for the 72 Base Interfaces, the 31 Derived Samples, and the Full Benchmark combining both. Higher is better for all metrics. \update{Green and red denote the best and worst performance, respectively, while orange indicates degraded performance after the repair rounds.} \zq{The \emph{with repair} rows report results for these models after up to three repair rounds} (\cref{sec:repair}).}
  \vspace{-2mm}
  \label{tab:overall-performance}
  \begin{tabular}{l*{13}{c}}
    \toprule
    \multirow{2}{*}{Model} 
      & \multicolumn{4}{c}{Base Interfaces} 
      & \multicolumn{4}{c}{Derived Samples} 
      & \multicolumn{5}{c}{Full Benchmark} \\
    \cmidrule(lr){2-5}
    \cmidrule(lr){6-9}
    \cmidrule(lr){10-14}
      & Exec.$\uparrow$ & Static$\uparrow$ & Data$\uparrow$ & Overall$\uparrow$
      & Exec.$\uparrow$ & Static$\uparrow$ & Data$\uparrow$ & Overall$\uparrow$
      & Exec.$\uparrow$ & Static$\uparrow$ & Data$\uparrow$ & Int.$\uparrow$ & Overall$\uparrow$ \\
    \midrule
    GLM 4.6V             & \bad{0.2361} & \bad{0.6722} & \bad{0.0090} & \bad{0.0529}
                         & \bad{0.3226} & \bad{0.6896} & \bad{0.0000}      & \bad{0.0741}
                         & \bad{0.2621} & \bad{0.6786} & \bad{0.0090} & \bad{0.0000}      & \bad{0.0592} \\
    Kimi K2.5            & \good{0.9444} & 0.7500 & 0.2004 & \good{0.3242}
                         & 0.9677 & 0.7646 & \good{0.3605} & 0.2737
                         & \good{0.9515} & 0.7545 & \good{0.2171} & \good{0.1168} & \good{0.3090} \\
    Qwen3.5 Plus         & 0.8056 & \good{0.7601} & 0.1917 & 0.2774
                         & 0.7079 & \good{0.7748} & 0.1247 & 0.2033
                         & 0.7767 & \good{0.7642} & 0.1781 & 0.0768 & 0.2551 \\
    Claude Sonnet 4.5    & 0.7083 & 0.7333 & \good{0.2041} & 0.2420
                         & \good{1.0000} & 0.7747 & 0.3027 & \good{0.2777}
                         & 0.7961 & 0.7489 & 0.2143 & 0.1077 & 0.2527 \\
    \update{GPT 5.4}             & 0.8194 & 0.6852 & 0.1051 & 0.2531
                         & 0.9677 & 0.7267 & 0.0506 & 0.2513
                         & 0.8641 & 0.6992 & 0.0887 & 0.0702 & 0.2525 \\
    \midrule
    \update{GLM 4.6V + Repair}            & 0.4167 & 0.7154 & 0.0497 & 0.1162
                         & 0.5161 & 0.7396 & 0.0352      & 0.1282
                         & 0.4466 & 0.7238 & 0.0453 & 0.0063      & 0.1198 \\
    \update{Kimi K2.5 + Repair}           & 0.9861 & 0.7523 & 0.3202 & 0.4044
                         & 0.9677 & 0.7708 & \worse{0.0894} & 0.2784
                         & 0.9806 & 0.7578 & 0.2508 & \worse{0.1056} & 0.3665 \\
    \update{Qwen3.5 Plus + Repair}        & 0.8056 & 0.7710 & 0.2396 & 0.3213
                         & 0.9355 & 0.7833 & \worse{0.0744} & 0.2691
                         & 0.8447 & 0.7751 & 0.1899 & 0.0769 & 0.3056 \\
    \update{Claude Sonnet 4.5 + Repair}   & 0.9444 & 0.7491 & 0.2873 & 0.3762
                         & 1.0000 & 0.7853 & \worse{0.0636} & 0.2829
                         & 0.9612 & 0.7604 & 0.2199 & \worse{0.0952} & 0.3481 \\
    \update{GPT 5.4 + Repair}             & 0.9167 & 0.6935 & 0.1200 & 0.2803
                         & 1.0000 & \worse{0.7201} & 0.0520 & 0.2574
                         & 0.9417 & 0.7020 & 0.0995 & \worse{0.0603} & 0.2734 \\
    \bottomrule
  \end{tabular}
  \vspace{-5mm}
\end{table*}

\begin{figure}[t!]
  \centering
  \includegraphics[width=\linewidth]{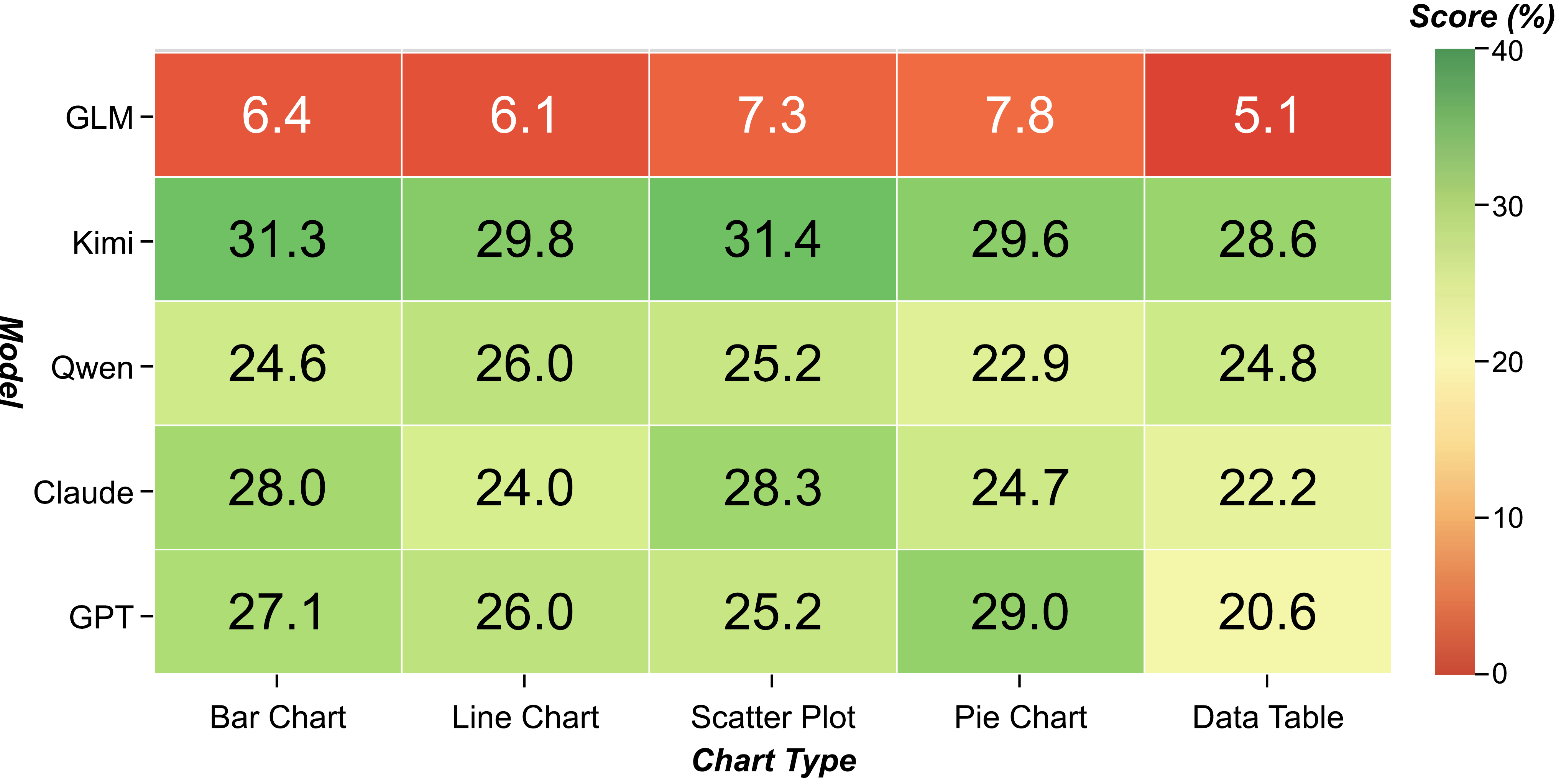}
  \caption{Performance across chart types (bar, line, scatter, pie, and table). Results are relatively consistent across types, with minor variation.}
\label{fig:bench-types}
\end{figure}


\section{Experimental Results}
Table~\ref{tab:overall-performance} summarizes the performance of \update{the five evaluated models} on \update{MV-Bench}. We report results on a representative 10\% subset, covering 72 Base Interfaces, 31 Derived Samples, and the Full Benchmark combining both partitions.

\subsection{Overall Performance Comparison}

\update{As shown in Table~\ref{tab:overall-performance}}, on the Full Benchmark, Kimi K2.5 attains the highest Overall score of 30.9\%, followed by Qwen3.5 Plus at 25.51\%\update{, with Claude Sonnet 4.5 (25.27\%) and GPT 5.4 (25.25\%) close behind}. The lowest Overall score is obtained by GLM 4.6V at 5.92\%. Across all \update{five} models, Exec. ranges from 26.21\% to 95.15\%, Static from 67.86\% to 76.42\%, Data from 0.90\% to 21.71\%, and Int. from 0.00\% to 11.68\%.
We observe two key patterns. First, all models achieve substantially higher Static scores than Data and Int. scores, with average gaps of 58.76 and 65.48 percentage points, respectively. This confirms that, \zq{under the single-pass generation setting we evaluate}, current MLLMs are more capable of replicating visual appearance than handling the data semantics and interactive logic beneath it. \zq{We test whether this gap is merely an artifact of the single-pass setting in \cref{sec:repair}.} Second, the relative ranking of models remains consistent across the Base and Derived partitions, suggesting that our recomposition strategy does not introduce systematic bias.

\subsection{Performance Across Chart Types}

Figure~\ref{fig:bench-types} breaks down Overall scores by chart type across the five chart types in \update{MV-Bench}. Performance is relatively consistent across types, ranging from 20.3\% (data tables) to 23.5\% (scatter plots and bar charts), with line charts in between (22.4\%); models handle common chart structures reasonably well, while more diverse layouts and tabular representations are slightly more difficult. Model-wise trends are stable across chart types: Kimi leads in all five categories (28.6\%--31.4\%); Qwen and Claude form a middle tier, with Qwen ahead on line charts and data tables and Claude ahead on scatter plots; GLM trails across all types (consistently below 8\%). The stability of this ranking across chart types indicates that performance differences are driven primarily by model capability rather than chart-specific difficulty.

\subsection{Visual \zq{Fidelity} vs. \zq{Functionality}}
\zq{Our results show that MLLMs} reproduce the visual appearance of an interface far better than its functionality: for the same interface, Static scores are much higher than both Data and Int.\ scores (Table~\ref{tab:overall-performance}, Full Benchmark).
\zq{The pattern} is universal: every model \zq{scores higher} on Static than on Data or Int. Static minus Int. ranges from 62.90 to 68.74 percentage points across models, and Static minus Data from 53.46 to 66.96. 
Notably, this is not simply an artifact of weaker models \zq{failing everywhere}. Kimi K2.5, the strongest model overall, \zq{scores} 75.45\% on Static but only 11.68\% on Int., and GLM 4.6V \zq{scores} 67.86\% on Static yet 0.00\% on Int. 
\zq{Inspecting the rendered outputs, we further find that those closely matching the reference appearance often still score near zero on Int. Therefore, the functional gap appears to be a general capability limitation, not one that scaling model strength alone will close.}

\subsection{\update{Effect of Iterative Repair}}
\label{sec:repair}
\zq{We now report the repair condition described in \cref{sec:setup}, applying up to three repair rounds to each model's single-pass output (\cref{sec:codegen}); the \emph{with repair} rows of Table~\ref{tab:overall-performance} report the resulting scores. 

Repair improves executability substantially (Exec. rises by 2.9 to 18.5 percentage points across the five models) but static fidelity only modestly (0.3 to 5.2 percentage points), suggesting that many single-pass failures are recoverable compile- or render-level errors rather than design errors. 
However, the functional gap remains: interaction completeness is almost the same for GLM 4.6V and Qwen3.5 Plus, and \emph{decreases} for the other three despite their large executability gains; data binding improves on Base Interfaces (1.5 to 12
percentage points) but degrades on Derived Samples for three of the five models. 
The Overall gains (2.1 to 9.5 percentage points) are thus driven almost entirely by executability, not by data semantics or interaction, and the regressions on Derived Samples suggest that repairs targeting compile and render errors can introduce new data-binding errors on the compositionally novel partition.}

\subsection{Qualitative Error Analysis}

\begin{figure*}[t!]
  \centering
  \includegraphics[width=0.99\linewidth]{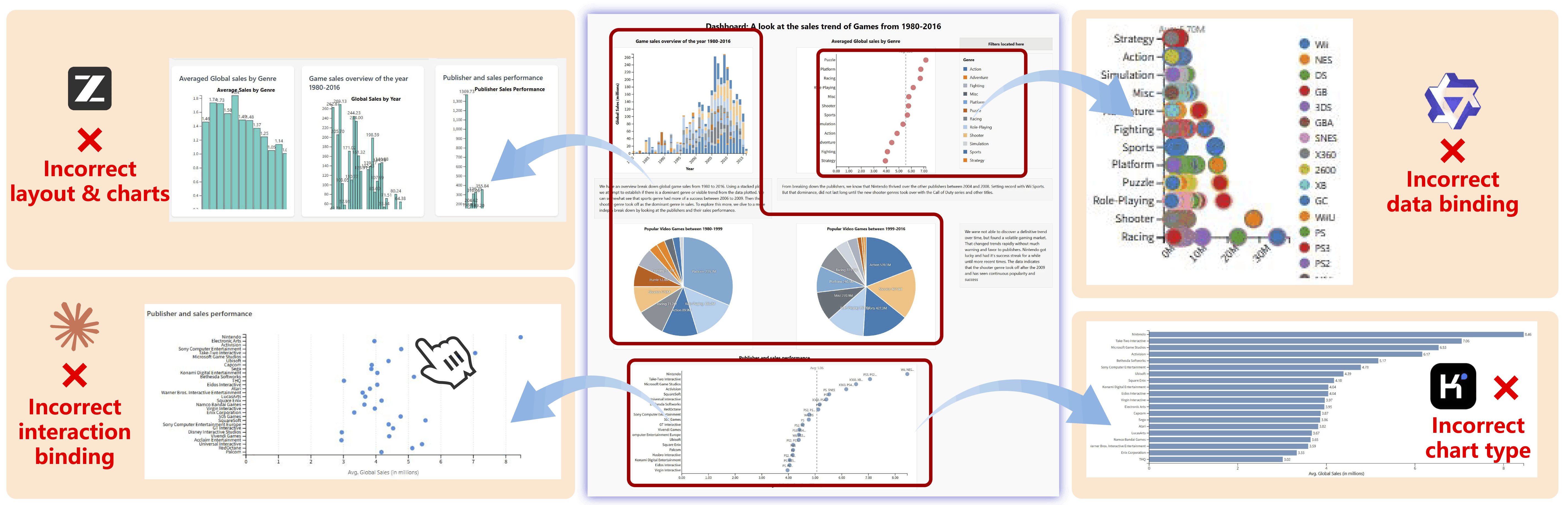}
  \vspace{-1mm}
  \caption{Representative failure modes of MLLM-generated interfaces, including incorrect layout and chart composition, incorrect data binding, incorrect interaction binding, and incorrect chart type selection.}
  \vspace{-5mm}
\label{fig:failure}
\end{figure*}

To better understand the failure modes underlying the quantitative results, we examine generated code and classify recurring errors into four categories. Figure~\ref{fig:failure} presents representative examples.

\noindent\textbf{Data binding errors} occur when the generated code uses fabricated or hardcoded data rather than the provided dataset---e.g., plausible-looking bars whose values correspond to no field in the source data. \textbf{Interaction omission} produces a static rendering with no interaction logic whatsoever, despite the ground truth requiring brushing, filtering, or other dynamic behavior. \textbf{Incorrect interaction binding} implements interaction but connects it to the wrong elements or data fields (e.g., a filter operating on the wrong dimension), suggesting models can recognize the need for interaction but struggle to reason about the correct data-semantic relationships. \textbf{Layout and structural errors}---misaligned views, incorrect aspect ratios, missing views, broken CSS/HTML---are less frequent but show that even at the visual level, constructing coordinated multi-view layouts remains challenging.

\subsection{Base Interfaces vs. Derived Samples}
We compare model performance between the 72 Base Interfaces and the 31 Derived Samples to validate the consistency of \update{MV-Bench}. The relative ordering of models is preserved across both partitions, and absolute score differences are small (within 3.67\% on average for Overall). Derived Samples yield slightly lower scores (21.60\% vs. 22.99\% Overall), which is expected since recomposition introduces novel chart-type and data-source combinations not present in the base set. This modest decrease confirms that derived samples provide meaningful additional evaluation coverage without fundamentally altering the difficulty distribution, and that our construction pipeline produces stable ground truth across the recomposition process.
\section{Discussion}
\label{sec:discussion}

\subsection{Key Findings and Implications}
Our evaluation reveals that, \zq{under the single-pass agentic-coding generation setting we study}, current MLLMs can reproduce the visual surface of \update{coordinated multi-view interfaces} but struggle with the data semantics and coordinated interaction logic that make them functional. We highlight \update{four} findings and discuss their underlying causes.

(1) \textbf{Visual replication outperforms functional correctness across all models.} The gap between Static scores and both Data and Int. scores is consistent across models, chart types, and benchmark partitions, suggesting \zq{a substantial limitation under this single-pass setting} rather than a dataset-specific artifact. We attribute this to a \textbf{representation asymmetry}: a dashboard's visual layout is directly observable in the input screenshot, but data bindings and interaction logic are latent properties that must be inferred from indirect cues such as axis labels, shared color encodings, and spatial proximity of views--and current vision encoders are optimized for capturing visual patterns, not for recovering these latent semantic structures.

(2) \textbf{Cross-view interaction is the hardest capability, with brushing and linking as the primary bottleneck.} Brushing and linking requires inferring shared data dimensions across views and implementing bidirectional state propagation, a form of cross-view reasoning that goes beyond visual pattern matching. We attribute this to a \textbf{training distribution gap}: existing MLLM training data predominantly consists of self-contained, single-component code generation tasks (individual charts, isolated UI widgets), whereas multi-view coordination requires reasoning about shared state, data flow between components, and the causal effects of user actions across view boundaries--capabilities rarely exercised in standard training schemes.

(3) \textbf{Data binding errors are pervasive but invisible to visual-only evaluation.} Our Data metric reveals that models frequently generate charts that look correct but bind wrong fields or fabricate values--undetectable under a purely visual evaluation protocol, underscoring the importance of evaluating data binding as an independent dimension. We attribute this to \textbf{shortcut learning}: models learn to produce visually plausible chart structures (correct axis layout, appropriate number of bars) without grounding them in the actual data, effectively hallucinating data-driven visualizations.

\zq{(4) \textbf{The functional gap is a capability limit, not a generation-budget artifact.} The patterns above persist under iterative repair: up to three repair rounds (\cref{sec:repair}) substantially raise executability but leave the Data and Int.\ gap largely intact, and on the harder Derived samples error-driven edits can even reduce data-binding correctness. We attribute this to a \textbf{feedback asymmetry}: compile and render errors are localized and directly fixable, whereas data-binding and interaction failures stem from a wrong inference about which fields and views are coordinated, which execution feedback can flag but not correct. The gap therefore holds under iterative, feedback-driven generation as well as single-pass generation; \cref{sec:repair} gives the full analysis.}

\update{More broadly, our findings suggest that visual fidelity is an unreliable proxy for functional correctness, so MLLM-based dashboard tools need to verify data binding and interaction by running the generated code, not by the appearance alone. MV-Bench's instances, sandboxed execution, and replayable interactions could serve this role directly, providing such feedback during iterative generation.}

\subsection{Limitations and Future Directions}
Our work has several limitations in source coverage, implementation, and evaluation, each pointing to promising future research directions.

(1) \textbf{Source coverage and generalizability.} \update{MV-Bench} is built from Tableau dashboards and reflects one authoring ecosystem's design patterns. Extending the pipeline to formats such as Apache Superset or Grafana would broaden coverage and test whether the coordination challenges we identify are universal or Tableau-specific, motivating \emph{specification-agnostic} intermediate representations for multi-view data dependencies across heterogeneous visual analytics systems.

(2) \textbf{Implementation space and training strategies.} Our fixed reference stack improves reproducibility but narrows the implementation space, reflecting a deeper limitation: current MLLMs treat each view as an independent generation target, lacking explicit representations of shared state and cross-view data flow. Addressing this may require \emph{interaction-aware training strategies} that expose models to the causal structure of coordinated interfaces during training.

(3) \textbf{Evaluation granularity.} Our metrics do not exhaustively capture all data-semantic errors, and interaction replay covers only the episodes encoded in the specification rather than all possible user trajectories. Beyond finer-grained data-semantic assessment, this motivates mechanisms that can localize where and why a model fails to implement correct coordination. \zq{We evaluate on three functional dimensions: static visual fidelity, data binding correctness, and interaction completeness. Properties such as code architecture, embedded analytical modeling, non-visualization UI elements, and multi-step analytical workflows are left outside this boundary, as they either resist reproducible automatic verification at our scale or fall outside the coordinated multi-view construction task we target. Finer decompositions of data-semantic and interaction correctness remain compatible with this three-way scheme and are a natural refinement for future work.}

(4) \textbf{Inference reproducibility.} We evaluate all models under low-stochasticity decoding (temperature=0) to minimize sampling-induced variance, but API-based LLM inference is not fully deterministic in practice (floating-point/scheduling effects, backend batching, service-side updates), and computational constraints prevent repeated runs for all samples. This may introduce run-to-run variance in absolute scores; we nevertheless observe consistent trend-level conclusions across independently constructed base and derived partitions, providing indirect evidence of result stability. Future work will conduct repeated evaluations on representative subsets and report dispersion statistics.

(5) \textbf{Data contamination.} Since the DMiner corpus is public on GitHub and the evaluated MLLMs are trained on large-scale web corpora, some benchmark instances may overlap with training data, which we cannot fully rule out without access to the models' training sets. Two observations mitigate this concern: our instances are translated web interfaces with generated code that did not exist prior to our pipeline rather than raw Tableau workbooks, and the consistently low interaction scores across all models suggest that even partially memorized visual patterns were not accompanied by memorized functional logic. Developing contamination detection methods for code generation benchmarks remains an important future direction.

\zq{(6) \textbf{Target-stack confound.} Our reference implementations target a fixed imperative stack (React, TypeScript, and D3), and so do the prompts and ground truth we evaluate against. This fixed stack requires models to manage shared state, DOM-level data bindings, and event-driven coordination explicitly. Such explicit management may make data binding and interaction harder than they would be under declarative grammars such as Vega-Lite, where views, selections, and cross-view links are expressed as compositional specifications rather than imperative code. 
Our results therefore reflect MLLM performance on this stack, not on dashboard construction in general. Whether the gap between visual fidelity and functional correctness persists under declarative targets remains untested, and retargeting MV-Bench's pipeline to such targets is left to future work.}

\zq{(7) \textbf{Chart and interaction vocabulary.} MV-Bench's five chart types and three coordination patterns, inherited from the DMiner corpus~\cite{Lin2023}, cover frequent forms of cross-view coordination but exclude more complex constructs. Examples include geographic and hierarchical visualizations, parameterized controls, drill-down navigation, compound interactions that combine several coordination mechanisms, and interfaces that require persistent application-level state across views. The current scope is therefore a verifiable subset of coordinated multi-view authoring rather than a complete one. Extending the construction pipeline to these richer vocabularies is a direct line of future work.}
\section{Conclusion}
We present \update{MV-Bench}, a benchmark for evaluating multimodal large language models (MLLMs) on \update{coordinated multi-view interface construction}. Compared to single-chart settings, this task requires models to generate code that integrates data semantics, multi-view coordination, and interactive behavior.
To support this evaluation, we develop a multi-stage pipeline that turns Tableau workbooks into executable \update{coordinated multi-view interfaces} while preserving their data bindings and interaction definitions. Using this pipeline, \zq{MV-Bench} provides 92 base interfaces covering five chart types and three interaction patterns, each with executable code, rendered output, datasets, and structured interaction annotations, \zq{which} we extend by systematic recomposition into 956 derived samples.
We evaluate \zq{five MLLMs, each queried as a coding agent in a single-pass setting.} While current models can often reproduce the visual layout of an interface, they score substantially lower on data binding correctness and interaction completeness, and \zq{this gap holds across models, chart types, and interaction patterns. Allowing up to three repair rounds with the same feedback used in our reference pipeline raises executability but does not consistently close the gap, indicating that it reflects a capability limit rather than an artifact of single-pass generation.}

\section*{Acknowledgments}
This study is supported by grants from NSFC (No. 62372271) and the Shandong Provincial Natural Science Foundation (No. ZR2026QC1575).

\bibliographystyle{abbrv-doi-hyperref}
\bibliography{template} 

@InProceedings{Radford2021,
  title = 	 {Learning Transferable Visual Models From Natural Language Supervision},
  author =       {Radford, Alec and Kim, Jong Wook and Hallacy, Chris and Ramesh, Aditya and Goh, Gabriel and Agarwal, Sandhini and Sastry, Girish and Askell, Amanda and Mishkin, Pamela and Clark, Jack and Krueger, Gretchen and Sutskever, Ilya},
  booktitle = 	 {Proceedings of the 38th International Conference on Machine Learning},
  pages = 	 {8748--8763},
  year = 	 {2021},
  volume = 	 {139},
  series = 	 {Proceedings of Machine Learning Research},
  publisher =    {PMLR}
}

@article{geminiteam2025,
author = {Anil, Rohan and Borgeaud, Sebastian and others},
title = {Gemini: A Family of Highly Capable Multimodal Models},
year = {2023},
journal = {arXiv preprint arXiv:2312.11805},
eprint = {2312.11805},
archivePrefix = {arXiv},
primaryClass = {cs.CL},
doi = {10.48550/arXiv.2312.11805},
url = {https://arxiv.org/abs/2312.11805}
}

@article{KimiTeam2025,
  title={Kimi K2.5: Visual Agentic Intelligence},
  author={Bai, Tongtong and Bai, Yifan and Bao, Yiping and Cai, SH and Cao, Yuan and Charles, Y and Che, HS and Chen, Cheng and Chen, Guanduo and others},
  journal={arXiv preprint arXiv:2602.02276},
  doi={10.48550/arXiv.2602.02276},
  year={2026}
}

@article{Qwen2023,
  title={Qwen Technical Report},
  author={Jinze Bai and Shuai Bai and Yunfei Chu and Zeyu Cui and Kai Dang and Xiaodong Deng and Yang Fan and Wenbin Ge and Yu Han and Fei Huang and others},
  journal={arXiv preprint arXiv:2309.16609},
  doi={10.48550/arXiv.2309.16609},
  year={2023}
}

@article{GLM2024,
  title={Chatglm: A family of large language models from glm-130b to glm-4 all tools},
  author={Zeng, Aohan and Xu, Bin and Wang, Bowen and Zhang, Chenhui and Yin, Da and Zhang, Dan and Rojas, Diego and Feng, Guanyu and Zhao, Hanlin and others},
  journal={arXiv preprint arXiv:2406.12793},
  doi={10.48550/arXiv.2406.12793},
  year={2024}
}

@article{Heer2012,
author = {Heer, Jeffrey and Shneiderman, Ben},
title = {Interactive dynamics for visual analysis},
year = {2012},
issue_date = {April 2012},
publisher = {Association for Computing Machinery},
address = {New York, NY, USA},
volume = {55},
number = {4},
issn = {0001-0782},
doi = {10.1145/2133806.2133821},
journal = {Commun. ACM},
pages = {45–54},
numpages = {10}
}

@book{Munzner2015,
  author = {Munzner, Tamara},
  description = {Visualization Analysis and Design},
  isbn = {9781498759717},
  publisher = {CRC Press},
  series = {AK Peters Visualization Series},
  title = {Visualization Analysis and Design},
  doi={10.1145/3721241.3733989},
  year = 2015
}

@article{Shen2022,
author = {Shen, Leixian and Shen, Enya and Luo, Yuyu and Yang, Xiaocong and Hu, Xuming and Zhang, Xiongshuai and Tai, Zhiwei and Wang, Jianmin},
title = {Towards Natural Language Interfaces for Data Visualization: A Survey},
year = {2023},
issue_date = {June 2023},
publisher = {IEEE Educational Activities Department},
address = {USA},
volume = {29},
number = {6},
issn = {1077-2626},
url = {https://doi.org/10.1109/TVCG.2022.3148007},
doi = {10.1109/TVCG.2022.3148007},
journal = {IEEE Transactions on Visualization and Computer Graphics},
pages = {3121–3144},
numpages = {24}
}

@INPROCEEDINGS{Vaquez2024,
  author={Vázquez, Pere-Pau},
  booktitle={2024 IEEE 17th Pacific Visualization Conference (PacificVis)}, 
  title={Are LLMs ready for Visualization?}, 
  year={2024},
  volume={},
  number={},
  pages={343-352},
  doi={10.1109/PacificVis60374.2024.00049}
}

@ARTICLE{Ren2019,
  author={Ren, Donghao and Lee, Bongshin and Brehmer, Matthew},
  journal={IEEE Transactions on Visualization and Computer Graphics}, 
  title={Charticulator: Interactive Construction of Bespoke Chart Layouts}, 
  year={2019},
  volume={25},
  number={1},
  pages={789-799},
  doi={10.1109/TVCG.2018.2865158}
}

@inproceedings{Ge2021,
author = {Ge, Tong and Lee, Bongshin and Wang, Yunhai},
title = {CAST: Authoring Data-Driven Chart Animations},
year = {2021},
isbn = {9781450380966},
publisher = {Association for Computing Machinery},
address = {New York, NY, USA},
url = {https://doi.org/10.1145/3411764.3445452},
doi = {10.1145/3411764.3445452},
booktitle = {Proceedings of the 2021 CHI Conference on Human Factors in Computing Systems},
articleno = {24},
numpages = {15},
series = {CHI '21}
}

@inproceedings{Liu2018,
author = {Liu, Zhicheng and Thompson, John and Wilson, Alan and Dontcheva, Mira and Delorey, James and Grigg, Sam and Kerr, Bernard and Stasko, John},
title = {Data Illustrator: Augmenting Vector Design Tools with Lazy Data Binding for Expressive Visualization Authoring},
year = {2018},
isbn = {9781450356206},
publisher = {Association for Computing Machinery},
address = {New York, NY, USA},
url = {https://doi.org/10.1145/3173574.3173697},
doi = {10.1145/3173574.3173697},
booktitle = {Proceedings of the 2018 CHI Conference on Human Factors in Computing Systems},
pages = {1–13},
numpages = {13},
location = {Montreal QC, Canada},
series = {CHI '18}
}

@inproceedings{Xie2024,
author = {Xie, Liwenhan and Zheng, Chengbo and Xia, Haijun and Qu, Huamin and Zhu-Tian, Chen},
title = {WaitGPT: Monitoring and Steering Conversational LLM Agent in Data Analysis with On-the-Fly Code Visualization},
year = {2024},
isbn = {9798400706288},
publisher = {Association for Computing Machinery},
address = {New York, NY, USA},
url = {https://doi.org/10.1145/3654777.3676374},
doi = {10.1145/3654777.3676374},
booktitle = {Proceedings of the 37th Annual ACM Symposium on User Interface Software and Technology},
articleno = {119},
numpages = {14},
location = {Pittsburgh, PA, USA},
series = {UIST '24}
}

@inproceedings{Niu2025,
    title = "{C}hart2{C}ode53: A Large-Scale Diverse and Complex Dataset for Enhancing Chart-to-Code Generation",
    author = "Niu, Tianhao  and
      Cui, Yiming  and
      Wang, Baoxin  and
      Xu, Xiao  and
      Yao, Xin  and
      Zhu, Qingfu  and
      Wu, Dayong  and
      Wang, Shijin  and
      Che, Wanxiang",
    booktitle = "Proceedings of the 2025 Conference on Empirical Methods in Natural Language Processing",
    year = "2025",
    address = "Suzhou, China",
    publisher = "Association for Computational Linguistics",
    url = "https://aclanthology.org/2025.emnlp-main.799/",
    doi = "10.18653/v1/2025.emnlp-main.799",
    pages = "15828--15844",
    ISBN = "979-8-89176-332-6"
}

@inproceedings{Dibia2023,
    title = "{LIDA}: A Tool for Automatic Generation of Grammar-Agnostic Visualizations and Infographics using Large Language Models",
    author = "Dibia, Victor",
    booktitle = "Proceedings of the 61st Annual Meeting of the Association for Computational Linguistics",
    year = "2023",
    address = "Toronto, Canada",
    publisher = "Association for Computational Linguistics",
    url = "https://aclanthology.org/2023.acl-demo.11/",
    doi = "10.18653/v1/2023.acl-demo.11",
    vol = {3},
    pages = "113--126"
}

@inproceedings{Goswami2025,
author = {Goswami, Kanika and Mathur, Puneet and Rossi, Ryan and Dernoncourt, Franck},
title = {PlotGen: Multi-Agent LLM-based Scientific Data Visualization via Multimodal Retrieval Feedback},
year = {2025},
isbn = {9798400713316},
publisher = {Association for Computing Machinery},
address = {New York, NY, USA},
url = {https://doi.org/10.1145/3701716.3716888},
doi = {10.1145/3701716.3716888},
booktitle = {Companion Proceedings of the ACM on Web Conference 2025},
pages = {1672–1676},
numpages = {5},
location = {Sydney NSW, Australia},
series = {WWW '25}
}

@article{RevisionWorkflowLLM2025,
  author={Moreira, Gustavo and Ferreira, Leonardo and Veiga, Carolina and Hosseini, Maryam and Miranda, Fabio},
  journal={IEEE Transactions on Visualization and Computer Graphics}, 
  title={Urbanite: A Dataflow-Based Framework for Human-AI Interactive Alignment in Urban Visual Analytics}, 
  year={2026},
  volume={32},
  number={1},
  pages={1065-1075},
  doi={10.1109/TVCG.2025.3634644}
}

@article{OpenAI2024a,
  title={{GPT-4o} System Card},
  author={{OpenAI}},
  journal={arXiv preprint arXiv:2410.21276},
  doi={10.48550/arXiv.2410.21276},
  year={2024}
}

@article{Anthropic2024a,
  title={Claude},
  author={{Anthropic}},
  year={2024},
  note={\url{https://www.anthropic.com/}}
}

@ARTICLE{Satyanarayan2017a,
  author={Satyanarayan, Arvind and Moritz, Dominik and Wongsuphasawat, Kanit and Heer, Jeffrey},
  journal={IEEE Transactions on Visualization and Computer Graphics}, 
  title={Vega-Lite: A Grammar of Interactive Graphics}, 
  year={2017},
  volume={23},
  number={1},
  pages={341-350},
  doi={10.1109/TVCG.2016.2599030}
}

@article{Bostock2011a,
  author={Bostock, Michael and Ogievetsky, Vadim and Heer, Jeffrey},
  journal={IEEE Transactions on Visualization and Computer Graphics}, 
  title={D3: Data-Driven Documents}, 
  year={2011},
  volume={17},
  number={12},
  pages={2301-2309},
  doi={10.1109/TVCG.2011.185}
}

@article{LI2018136a,
title = {ECharts: A declarative framework for rapid construction of web-based visualization},
journal = {Visual Informatics},
volume = {2},
number = {2},
pages = {136-146},
year = {2018},
issn = {2468-502X},
doi = {https://doi.org/10.1016/j.visinf.2018.04.011},
url = {https://www.sciencedirect.com/science/article/pii/S2468502X18300068},
author = {Deqing Li and Honghui Mei and Yi Shen and Shuang Su and Wenli Zhang and Junting Wang and Ming Zu and Wei Chen}
}

@book{Sievert2020a,
author = {Sievert, Carson},
title = {Interactive Web-Based Data Visualization with {R}, Plotly, and Shiny},
year = {2020},
publisher = {Chapman \& Hall/CRC},
isbn = {9781138331457},
doi = {10.1201/9780429447273},
url = {https://www.routledge.com/Interactive-Web-Based-Data-Visualization-with-R-Plotly-and-Shiny/Sievert/p/book/9781138331457}
}

@article{Luo2022a,
  author={Luo, Yuyu and Tang, Nan and Li, Guoliang and Tang, Jiawei and Chai, Chengliang and Qin, Xuedi},
  journal={IEEE Transactions on Visualization and Computer Graphics}, 
  title={Natural Language to Visualization by Neural Machine Translation}, 
  year={2022},
  volume={28},
  number={1},
  pages={217-226},
  doi={10.1109/TVCG.2021.3114848}
}

@inproceedings{Luo2021a,
author = {Luo, Yuyu and Tang, Nan and Li, Guoliang and Chai, Chengliang and Li, Wenbo and Qin, Xuedi},
title = {Synthesizing Natural Language to Visualization (NL2VIS) Benchmarks from NL2SQL Benchmarks},
year = {2021},
isbn = {9781450383431},
publisher = {Association for Computing Machinery},
address = {New York, NY, USA},
url = {https://doi.org/10.1145/3448016.3457261},
doi = {10.1145/3448016.3457261},
booktitle = {Proceedings of the 2021 International Conference on Management of Data},
pages = {1235–1247},
numpages = {13},
location = {Virtual Event, China},
series = {SIGMOD '21}
}

@article{Tian2025a,
  author={Tian, Yuan and Cui, Weiwei and Deng, Dazhen and Yi, Xinjing and Yang, Yurun and Zhang, Haidong and Wu, Yingcai},
  journal={IEEE Transactions on Visualization and Computer Graphics}, 
  title={ChartGPT: Leveraging LLMs to Generate Charts From Abstract Natural Language}, 
  year={2025},
  volume={31},
  number={3},
  pages={1731-1745},
  doi={10.1109/TVCG.2024.3368621}
}

@article{Wang2024a,
  author={Wang, Chenglong and Thompson, John and Lee, Bongshin},
  journal={IEEE Transactions on Visualization and Computer Graphics}, 
  title={Data Formulator: AI-Powered Concept-Driven Visualization Authoring}, 
  year={2024},
  volume={30},
  number={1},
  pages={1128-1138},
  doi={10.1109/TVCG.2023.3326585}
}

@ARTICLE{Zhao2025a,
  author={Zhao, Yuheng and Zhang, Yixing and Zhang, Yu and Zhao, Xinyi and Wang, Junjie and Shao, Zekai and Turkay, Cagatay and Chen, Siming},
  journal={IEEE Transactions on Visualization and Computer Graphics}, 
  title={LEVA: Using Large Language Models to Enhance Visual Analytics}, 
  year={2025},
  volume={31},
  number={3},
  pages={1830-1847},
  doi={10.1109/TVCG.2024.3368060}}

@ARTICLE{Zhao2025b,
  author={Zhao, Yuheng and Wang, Junjie and Xiang, Linbing and Zhang, Xiaowen and Guo, Zifei and Turkay, Cagatay and Zhang, Yu and Chen, Siming},
  journal={IEEE Transactions on Visualization and Computer Graphics}, 
  title={LightVA: Lightweight Visual Analytics With LLM Agent-Based Task Planning and Execution}, 
  year={2025},
  volume={31},
  number={9},
  pages={6162-6177},
  doi={10.1109/TVCG.2024.3496112}}

@inproceedings{Beltramelli2017a,
  author = {Beltramelli, Tony},
  title = {pix2code: Generating Code from a Graphical User Interface Screenshot},
  year = {2018},
  isbn = {9781450358972},
  publisher = {Association for Computing Machinery},
  address = {New York, NY, USA},
  url = {https://doi.org/10.1145/3220134.3220135},
  doi = {10.1145/3220134.3220135},
  booktitle = {Proceedings of the ACM SIGCHI Symposium on Engineering Interactive Computing Systems},
  articleno = {3},
  numpages = {6},
  location = {Paris, France},
  series = {EICS '18}
}

@inproceedings{Deka2017a,
    author = {Deka, Biplab and Huang, Zifeng and Franzen, Chad and Hibschman, Joshua and Afergan, Daniel and Li, Yang and Nichols, Jeffrey and Kumar, Ranjitha},
    title = {Rico: A Mobile App Dataset for Building Data-Driven Design Applications},
    year = {2017},
    isbn = {9781450349819},
    publisher = {Association for Computing Machinery},
    address = {New York, NY, USA},
    url = {https://doi.org/10.1145/3126594.3126651},
    doi = {10.1145/3126594.3126651},
    booktitle = {Proceedings of the 30th Annual ACM Symposium on User Interface Software and Technology},
    pages = {845–854},
    numpages = {10},
    location = {Qu\'{e}bec City, QC, Canada},
    series = {UIST '17}
}

@article{Laurencon2024a,
  title={Unlocking the Conversion of Web Screenshots into {HTML} Code with the {WebSight} Dataset},
  author={Lauren\c{c}on, Hugo and Tronchon, L\'{e}o and Sanh, Victor},
  journal={arXiv preprint arXiv:2403.09029},
  doi={10.48550/arXiv.2403.09029},
  year={2024}
}

@inproceedings{Li2024b,
    title = "{S}ketch2{C}ode: Evaluating Vision-Language Models for Interactive Web Design Prototyping",
    author = "Li, Ryan  and
      Zhang, Yanzhe  and
      Yang, Diyi",
    booktitle = "Proceedings of the 2025 Conference of the Nations of the Americas Chapter of the Association for Computational Linguistics: Human Language Technologies",
    year = "2025",
    address = "Albuquerque, New Mexico",
    publisher = "Association for Computational Linguistics",
    url = "https://aclanthology.org/2025.naacl-long.198/",
    doi = "10.18653/v1/2025.naacl-long.198",
    volume = {1},
    pages = "3921--3955",
    ISBN = "979-8-89176-189-6"
}

@article{Xie2025a,
  title={Widget2Code: From Visual Widgets to UI Code via Multimodal LLMs},
  author={Zhang, Houston H and Zhang, Tao and Lin, Baoze and Xue, Yuanqi and Zhu, Yincheng and Liu, Huan and Gu, Li and Ye, Linfeng and Wang, Ziqiang and Zuo, Xinxin and others},
  journal={arXiv preprint arXiv:2512.19918},
  doi={10.48550/arXiv.2512.19918},
  year={2025}
}

@inproceedings{Shi2025,
author = {Shi, Shengze and Ren, Tao and Zhu, Guoliang and Feng, Guandong and Hu, Jun},
title = {Closing the Feedback Loop in Text2Vis: Refining Visualization with Vision-Language Models},
year = {2025},
isbn = {9798400720352},
publisher = {Association for Computing Machinery},
address = {New York, NY, USA},
url = {https://doi.org/10.1145/3746027.3755862},
doi = {10.1145/3746027.3755862},
booktitle = {Proceedings of the 33rd ACM International Conference on Multimedia},
pages = {9053–9061},
numpages = {9},
location = {Dublin, Ireland},
series = {MM '25}
}

@inproceedings{Si2024a,
    title = "{D}esign2{C}ode: Benchmarking Multimodal Code Generation for Automated Front-End Engineering",
    author = "Si, Chenglei  and
      Zhang, Yanzhe  and
      Li, Ryan  and
      Yang, Zhengyuan  and
      Liu, Ruibo  and
      Yang, Diyi",
    booktitle = "Proceedings of the 2025 Conference of the Nations of the Americas Chapter of the Association for Computational Linguistics: Human Language Technologies",
    year = "2025",
    address = "Albuquerque, New Mexico",
    publisher = "Association for Computational Linguistics",
    volume={1},
    url = "https://aclanthology.org/2025.naacl-long.199/",
    doi = "10.18653/v1/2025.naacl-long.199",
    pages = "3956--3974"
}

@article{Gao2024a,
  title={A Rule-Based Approach for {UI} Migration from {Android} to {iOS}},
  author={Gao, Yi and Hu, Xing and Peng, Tongtong and Jiang, He},
  journal={arXiv preprint arXiv:2409.16656},
  doi={10.48550/arXiv.2409.16656},
  year={2024}
}

@inproceedings{Gong2024a,
    author = {Gong, Lina and Wang, Chen and Cui, Di and Huang, Yujun and Wei, Mingqiang},
    title = {UITrans: Seamless UI Translation from Android to HarmonyOS},
    year = {2025},
    isbn = {9798400719264},
    publisher = {Association for Computing Machinery},
    address = {New York, NY, USA},
    url = {https://doi.org/10.1145/3755881.3755933},
    doi = {10.1145/3755881.3755933},
    booktitle = {Proceedings of the 16th International Conference on Internetware},
    pages = {142–146},
    numpages = {5},
    series = {Internetware '25}
}

@Inbook{Keim2008a,
    author="Keim, Daniel
    and Andrienko, Gennady
    and Fekete, Jean-Daniel
    and G{\"o}rg, Carsten
    and Kohlhammer, J{\"o}rn
    and Melan{\c{c}}on, Guy",
    title="Visual Analytics: Definition, Process, and Challenges",
    bookTitle="Information Visualization: Human-Centered Issues and Perspectives",
    year="2008",
    publisher="Springer Berlin Heidelberg",
    address="Berlin, Heidelberg",
    pages="154--175",
    isbn="978-3-540-70956-5",
    doi="10.1007/978-3-540-70956-5_7",
    url="https://doi.org/10.1007/978-3-540-70956-5_7"
}

@inproceedings{Roberts2007a,
  author={Roberts, Jonathan C.},
  booktitle={Fifth International Conference on Coordinated and Multiple Views in Exploratory Visualization (CMV 2007)}, 
  title={State of the Art: Coordinated \& Multiple Views in Exploratory Visualization}, 
  year={2007},
  volume={},
  number={},
  pages={61-71},
  doi={10.1109/CMV.2007.20}
}

@book{Ward2010a,
author = {Ward, Matthew O. and Grinstein, Georges and Keim, Daniel A.},
title = {Interactive Data Visualization: Foundations, Techniques, and Applications},
year = {2010},
publisher = {A K Peters / CRC Press},
address = {Boca Raton, FL, USA},
isbn = {9781439865545},
doi = {10.1201/9780429108433},
url = {https://www.routledge.com/Interactive-Data-Visualization-Foundations-Techniques-and-Applications/Ward-Grinstein-Keim/p/book/9781439865545}
}

@article{Lin2023,
  author={Lin, Yanna and Li, Haotian and Wu, Aoyu and Wang, Yong and Qu, Huamin},
  journal={IEEE Transactions on Visualization and Computer Graphics}, 
  title={DMiner: Dashboard Design Mining and Recommendation}, 
  year={2024},
  volume={30},
  number={7},
  pages={4108-4121},
  doi={10.1109/TVCG.2023.3251344}
}

@article{Chen2025a,
  author={Chen, Nan and Zhang, Yuge and Xu, Jiahang and Ren, Kan and Yang, Yuqing},
  journal={IEEE Transactions on Visualization and Computer Graphics}, 
  title={VisEval: A Benchmark for Data Visualization in the Era of Large Language Models}, 
  year={2025},
  volume={31},
  number={1},
  pages={1301-1311},
  doi={10.1109/TVCG.2024.3456320}
}

@article{Wang2025a,
  author={Wang, Huichen Will and Gordon, Mitchell and Battle, Leilani and Heer, Jeffrey},
  journal={IEEE Transactions on Visualization and Computer Graphics}, 
  title={DracoGPT: Extracting Visualization Design Preferences from Large Language Models}, 
  year={2025},
  volume={31},
  number={1},
  pages={710-720},
  doi={10.1109/TVCG.2024.3456350}
}

@article{Jimenez2024a,
  title={{SWE-bench}: Can Language Models Resolve Real-World {GitHub} Issues?},
  author={Jimenez, Carlos E. and Yang, John and Wettig, Alexander and Yao, Shunyu and Pei, Kexin and Press, Ofir and Narasimhan, Karthik},
  journal={arXiv preprint arXiv:2310.06770},
  doi={10.48550/arXiv.2310.06770},
  year={2024}
}

@inproceedings{Masry2022a,
    title = "{C}hart{QA}: A Benchmark for Question Answering about Charts with Visual and Logical Reasoning",
    author = "Masry, Ahmed  and
      Long, Do Xuan  and
      Tan, Jia Qing  and
      Joty, Shafiq  and
      Hoque, Enamul",
    booktitle = "Findings of the Association for Computational Linguistics: ACL 2022",
    year = "2022",
    address = "Dublin, Ireland",
    publisher = "Association for Computational Linguistics",
    url = "https://aclanthology.org/2022.findings-acl.177/",
    doi = "10.18653/v1/2022.findings-acl.177",
    pages = "2263--2279"
}

@article{Xu2024a,
  title={{ChartBench}: A Benchmark for Complex Visual Reasoning in Charts},
  author={Xu, Zhengzhuo and others},
  journal={arXiv preprint arXiv:2312.15915},
  year={2024},
  doi={10.48550/arXiv.2312.15915}
}

@inproceedings{Yang2024a,
    title = "{M}at{P}lot{A}gent: Method and Evaluation for {LLM}-Based Agentic Scientific Data Visualization",
    author = "Yang, Zhiyu  and
      Zhou, Zihan  and
      Wang, Shuo  and
      Cong, Xin  and
      Han, Xu  and
      Yan, Yukun  and
      Liu, Zhenghao  and
      Tan, Zhixing  and
      Liu, Pengyuan  and
      Yu, Dong  and
      Liu, Zhiyuan  and
      Shi, Xiaodong  and
      Sun, Maosong",
    booktitle = "Findings of the Association for Computational Linguistics: ACL 2024",
    year = "2024",
    address = "Bangkok, Thailand",
    url = "https://aclanthology.org/2024.findings-acl.701/",
    doi = "10.18653/v1/2024.findings-acl.701",
    pages = "11789--11804",
}

@inproceedings{Chen2021a,
    title = "{P}lot{C}oder: Hierarchical Decoding for Synthesizing Visualization Code in Programmatic Context",
    author = "Chen, Xinyun  and
      Gong, Linyuan  and
      Cheung, Alvin  and
      Song, Dawn",
    booktitle = {Proceedings of the 59th Annual Meeting of the Association for Computational Linguistics and the 11th International Joint Conference on Natural Language Processing},
    volume = {1},
    year = "2021",
    address = "Online",
    publisher = "Association for Computational Linguistics",
    url = "https://aclanthology.org/2021.acl-long.169/",
    doi = "10.18653/v1/2021.acl-long.169",
    pages = "2169--2181"
}

@article{xia2025chartx,
  title={Chartx \& chartvlm: A versatile benchmark and foundation model for complicated chart reasoning},
  author={Xia, R. and Ye, H. and Yan, X. and others},
  journal={IEEE Transactions on Image Processing},
  year={2025}
}

@article{Han2023,
  title={Chartllama: A multimodal llm for chart understanding and generation},
  author={Han, Y. and Zhang, C. and Chen, X. and others},
  journal={arXiv preprint arXiv:2311.16483},
  doi={10.48550/arXiv.2311.16483},
  year={2023}
}

@article{wang2025chartoptimiser,
  title={ChartOptimiser: Task-driven Optimisation of Chart Designs},
  author={Wang, Y. and Pan, J. and Shi, D. and others},
  journal={arXiv preprint arXiv:2504.10180},
  doi={10.48550/arXiv.2504.10180},
  year={2025}
}

@article{Li2024a,
  author  = {Shuaimin Li and Xuanang Chen and Yuanfeng Song and Yunze Song and Chen Jason Zhang and Fei Hao and Lei Chen},
  title   = {Prompt4Vis: Prompting Large Language Models with Example Mining for Tabular Data Visualization},
  journal = {The VLDB Journal},
  year    = {2025},
  volume  = {34},
  number  = {4},
  pages   = {38},
  doi     = {10.1007/s00778-025-00912-0},
  url     = {https://doi.org/10.1007/s00778-025-00912-0}
}

@article{Chen2022a,
  title={Nl2interface: Interactive visualization interface generation from natural language queries},
  author={Chen, Yiru and Li, Ryan and Mac, Austin and Xie, Tianbao and Yu, Tao and Wu, Eugene},
  journal={IEEE Visualization Conference NLVIZ Workshop 2022},
  doi ={10.48550/arXiv.2209.08834}
}

@inproceedings{zhao2025libra,
author = {Zhao, Yue and Wang, Yunhai and Luo, Xu and Wang, Yanyan and Fekete, Jean-Daniel},
title = {Libra: An Interaction Model for Data Visualization},
year = {2025},
isbn = {9798400713941},
publisher = {Association for Computing Machinery},
address = {New York, NY, USA},
url = {https://doi.org/10.1145/3706598.3713769},
doi = {10.1145/3706598.3713769},
booktitle = {Proceedings of the 2025 CHI Conference on Human Factors in Computing Systems},
articleno = {1169},
numpages = {17},
series = {CHI '25}
}

\end{document}